\documentclass[journal]{IEEEtran}
\usepackage{graphicx}
\usepackage{float}
\usepackage{subfigure}

\usepackage{amssymb}
\usepackage{array}
\usepackage{amsmath}
\usepackage{url}
\usepackage{multirow}
\usepackage{booktabs}
\usepackage{threeparttable}
\usepackage{fixltx2e}
\usepackage{float}
\usepackage{booktabs}
\usepackage{graphicx}
\usepackage{diagbox}
\usepackage{mathtools}
\usepackage{amsmath}
\usepackage{bbding}
\usepackage{pifont}
\usepackage{wasysym}
\usepackage{tikz}
\usepackage{calc}
\usepackage{amssymb}
\usepackage{multirow}
\usepackage{rotating}
\usepackage{colortbl}
\usepackage{color}
\usepackage{ragged2e}
\usepackage[colorlinks,linkcolor=red,anchorcolor=blue,citecolor=green]{hyperref}
\usepackage{caption}
\usepackage{cite}
\usepackage{gensymb}
\usepackage{makecell}

\begin{document}
\title{Full Point Encoding for Local Feature Aggregation in 3D Point Clouds}
\author{Yong~He,
        Hongshan~Yu,
        Zhengeng~Yang,
        Xiaoyan Liu,
        Wei Sun,
        Ajmal~Mian 
\thanks{}
\thanks{Yong He, Hongshan Yu, Zhengeng Yang, Xiaoyan Liu, Wei Sun are with the National Engineering Laboratory for Robot Visual Perception and Control Technology, College of Electrical and Information Engineering, Hunan University, Lushan South Rd., Yuelu Dist., 410082, Changsha, China. This work was partially supported by the National Natural Science Foundation of China (Grants U2013203, 61973106).}
\thanks{Ajmal~Mian is with the Department of Computer Science, The University of Western Australia, WA 6009, Australia. He is the recipient of an Australian Research Council Future Fellowship Award (project number FT210100268) funded by the Australian Government.}}

\markboth{IEEE Transactions on Neural Networks and Learning Systems, Class Files,~Vol.~14, No.~8, August~2021}%
{Shell \MakeLowercase{\textit{et al.}}: Bare Demo of IEEEtran.cls for IEEE Journals}

\maketitle
\begin{abstract}
Point cloud processing methods exploit local point features and global context through aggregation which does not explicity model the internal correlations between local and global features. To address this problem, we propose full point encoding which is applicable to convolution and transformer architectures. Specifically, we propose Full Point Convolution (FPConv) and Full Point Transformer (FPTransformer) architectures. The key idea is to adaptively learn the weights from local and global geometric connections, where the connections are established through local and global correlation functions respectively. FPConv and FPTransformer simultaneously model the local and global geometric relationships as well as their internal correlations, demonstrating strong generalization ability and high performance. FPConv is incorporated in classical hierarchical network architectures to achieve local and global shape-aware learning. In FPTransformer, we introduce full point position encoding in self-attention, that hierarchically encodes each point position in the global and local receptive field. We also propose a shape aware downsampling block which takes into account the local shape and the global context. Experimental comparison to existing methods on benchmark datasets show the efficacy of FPConv and FPTransformer for semantic segmentation, object detection, classification, and normal estimation tasks. In particular, we achieve state-of-the-art semantic segmentation results of 76\% mIoU on S3DIS 6-fold and 72.2\% on S3DIS Area 5. 

\end{abstract}

\section{Introduction}
Point cloud processing has drawn considerable research interest due to its wide range of applications in autonomous driving\cite{chen2020boost,gao2023spatio,shu2023hierarchical}, robotics\cite{du2022novel}, and industrial automation\cite{khalid2019deep}. Learning effective features from raw point clouds is difficult due to its irregular nature. Early methods transformed points into regular grids (e.g. multi-view images\cite{lawin2017deep,boulch2018snapnet}, voxels\cite{tchapmi2017segcloud, maturana2015voxnet}), for seamless application of grid convolutions. However, the discretization process inevitably sacrifices important geometric information, distorts object shapes and results in a huge computational overhead.

\begin{figure}[!t]
\centering
\includegraphics[width= 0.95\columnwidth]{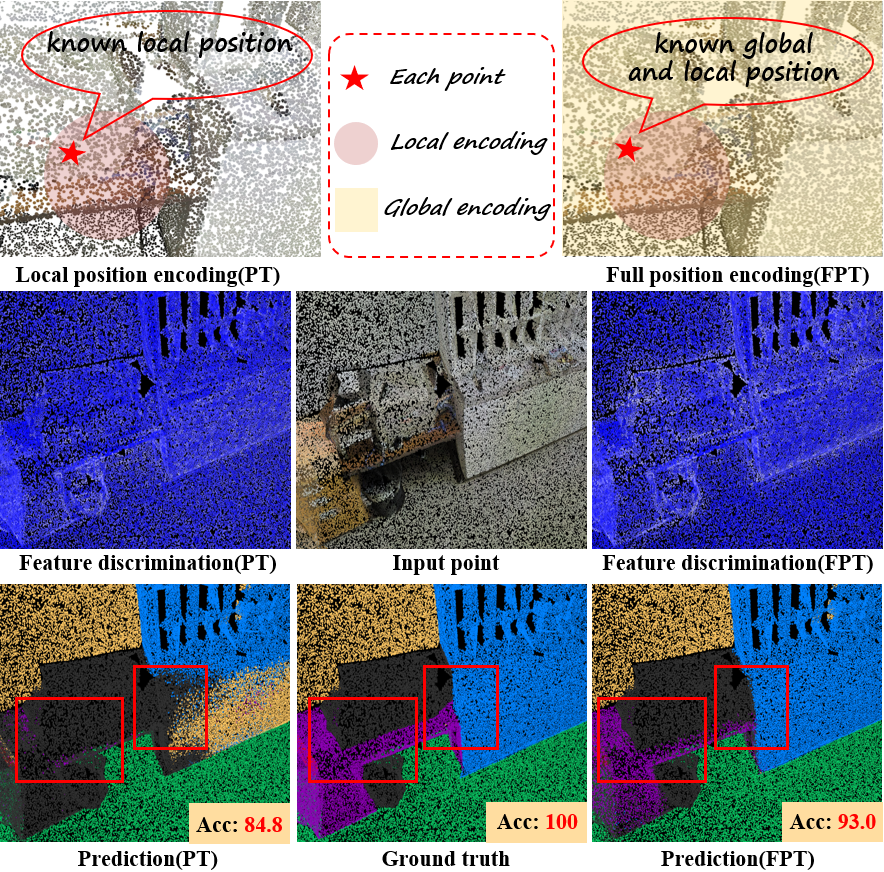}
\vspace{-2mm}
\caption{Full point position encoding makes each point aware of its position in the global and local receptive field (top), helping the proposed FPTransformer to simultaneously model global and local geometric cues 
as well as their internal correlations. FPTransformer enhances discriminating features of long-range shapes, e.g. edges, improving the full shape awareness to achieve better semantic segmentation (see inside red boxes). 
} 
\label{fig:teaser}
\vspace{-6mm}
\end{figure}

\indent To learn features from raw point clouds, the pioneering work PointNet\cite{qi2017pointnet} employs shared multi-layer perceptrons (MLPs) on each point and uses maxpooling to aggregate the features into a global representation. Such a design ignores local structures that are crucial for shape representation. To alleviate this problem, PointNet++\cite{qi2017pointnet++} additionally exploits local aggregation and adopts MLPs to learn local features. However, the local aggregator still treats the points independently, losing sight of the overall shape. 

\indent Inspired by 2D convolutions, some methods exploit point convolutions that learn the convolutional weights from local geometric connections. Early point convolutions use MLPs\cite{li2018pointcnn,simonovsky2017dynamic,wang2018deep,wu2019pointconv,hermosilla2018monte} as the convolutional weight function to learn weights from point coordinates. Other works approximate the weight functions as correlation functions\cite{shen2018mining,thomas2019kpconv,fey2018splinecnn,xu2018spidercnn}. 
Some methods associate coefficients (derived from point coordinates) \cite{wu2019pointconv,hermosilla2018monte,Lei_2020_CVPR} with weight functions to adjust the learned weights. 

\indent Self-attention was introduced to focus on important details in point clouds. Early methods in this category usually operate on global receptive field\cite{xie2018attentional,liu2019point2sequence,yang2019modeling,lee2019set,feng2020point,yan2020pointasnl} and learn the attention maps through a scalar dot-product. Differently, point transformer\cite{zhao2021point} employs self-attention to local receptive field and uses vector attention instead of scalar dot-product for learning the local geometric connections. Applying aggregation modules on local regions can effectively learn the local structure, but falls short of capturing the global context. To learn both local and global features, some methods integrate local and non-local feature aggregation modules into their network in a serial \cite{feng2020point,cheng2021net} or parallel \cite{yan2020pointasnl} manner. However, such pipelines still ignore the internal correlations between the local and global features besides significantly increasing the network parameters and leading to poor generalization.

 We propose full point position encoding, applicable to convolution and transformer architectures. Specifically, we propose Full Point Convolution (FPConv) and Full Point Transformer (FPTransformer) that exploit the local features and global context along with their interactions.  FPConv learns the local points layout in global receptive field by a global correlation function, and then learns the local points layout in local receptive field by a local correlation function, in a hierarchical manner. In FPTransformer, we introduce a novel position encoding scheme into the self attention that hierarchically encodes each point at the global and local level. This makes each point aware of its position in the global and local structure. Fig.~\ref{fig:teaser} intuitively compares the local position encoding in Point Transformer\cite{zhao2021point} to full point position encoding in our FPTransformer. FPTransformer enhances the feature discrimination of long-range shapes (e.g. edges), improving the full shape awareness. In FPTransformer, we also propose a shape aware downsampling block which takes into account the local shape and the global context. Our contributions are summarized as follows:

\begin{itemize}
\item We derive a general formulation for local feature aggregation methods, including local point-wise MLP, point convolution and point transformer, to highlight their limitations.

\vspace{1mm}

\item Based on the above general formulation, we propose a full point encoding method so as to simultaneously model local and global geometric features of point clouds along with their internal correlations. Using our novel encoding, we propose two network architectures, namely Full Point Convolution (FPConv) and Full Point Transformer (FPTransformer) and show promising results with both.

\vspace{1mm}
\item We propose a learnable downsampling block that performs local and global shape aware downsampling by incorporating the full point position encoding of the proposed FPTransformer into point-wise MLPs.
\end{itemize}
\indent 

We conduct extensive experiments on benchmark datasets to show the efficacy of our proposed methods and their strong generalization ability to different tasks such as semantic segmentation, object detection, classification and normal estimation. FPConv achieves  competitive results on semantic segmentation, classification, and normal estimation tasks compared to various point convolution-based methods. FPTransformer achieves state-of-the-art semantic segmentation performance with 76.0\% mIoU on S3DIS 6-fold and 72.2\% mIoU on S3DIS Area 5.
Incorporating FPTransformer into existing detection networks gives considerable performance improvement on ScanNetv2 and KITTI dataset.

\section{Related Work}\label{section2}
\noindent \textbf{Point-wise MLPs:} To maximally preserve the geometric information, recent deep neural networks prefer to directly process raw point clouds. The pioneering work PointNet\cite{qi2017pointnet} uses shared MLPs to exploit point-wise features and adopts a symmetric function (i.e. maxpooling) to aggregate these features into global features. However, it fails to consider the geometric relationships of local points. PointNet++\cite{qi2017pointnet++} addresses this issue by adopting a hierarchical network that benefits from efficient sampling\cite{hermosilla2018monte,hu2020randla,yan2020pointasnl,yang2019modeling,groh2018flex} and grouping\cite{qi2017pointnet++,li2018so,feng2020point,engelmann2018know,engelmann2017exploring,zhang2019shellnet} of the local region definition. However, MLP still treats each local point individually and ignores their geometric connections.

\indent Follow up works construct geometric connections between points to enrich point-wise features and then apply shared MLPs on them. For instance, some methods\cite{ran2022surface,ma2021rethinking} hand craft the geometric connections through curves, triangles, umbrella orientation or affine transformation. Graphs are also used to connect local points\cite{zhao2019pointweb,feng2020point,jiang2019hierarchical} or global points \cite{xu2021learning,wang2019dynamic,klokov2017escape,xu2020geometry}, for subsequent geometric representation i.e. edge, contour, curvature, and connectivity. Although geometric connections do not largely increase the learnable network parameters, the parameters of these hand-crafted representations or graphs must be optimized for different datasets with varying density or shape style.

\noindent \textbf{Point Convolution:} Inspired by 2D convolutions, various works successfully proposed novel  convolutions on points or their graphs that dynamically learn convolutional weights through functions that operate on the local geometric connections. Hence, the weight functions enable convolutions to be aware of the overall object shape. Early methods paid more attention to the weight function design. Most convolutional weight functions are approximated by MLP\cite{hermosilla2018monte,li2018pointcnn,wu2019pointconv,xu2021paconv,liu2019relation}. Other approaches treat weight functions as local correlation functions such as spline function\cite{fey2018splinecnn}, a family of polynomial functions\cite{xu2018spidercnn}, or standard unparameterized Fourier function\cite{wang2018local} to learn the convolution weights from the local geometric connections.

\indent Unlike the above dynamic convolution kernels, KPConv\cite{thomas2019kpconv} and KCNet\cite{shen2018mining} fixed the convolution kernel 
for robustness to varying point density. These networks predefine the kernel points on local region and learn convolutional weights on the kernel points from their geometric connections to local points using linear and Gaussian correlation functions, respectively. Here, the number and position of kernel points need be optimized for different datasets.

\indent Convolutional weights are generally learned from local geometric connections by weight functions. Hence, the convolutional weight learning highly depends on geometric connections. Some works construct additional low-level geometric connections (e.g. relative positions, 3D Euclidean distances\cite{liu2019relation,xu2021paconv}) to enrich the input to the weight function. Another line of works \cite{hermosilla2018monte,wu2019pointconv,Lei_2020_CVPR} associate a coefficient (derived from point coordinates) with the weight function to adjust the learned convolutional weights. Convolutional weights learned from low-level geometric connections cannot embed the global context in the convolution operation. Besides, point cloud resolution decreases at deeper layers in many networks and the geometric connections constructed by the sparse points may get distorted, leading to non-robust weights learned by the weight function.

\noindent \textbf{Point Transformer:} Early self attention modules operate on global points\cite{xie2018attentional,liu2019point2sequence,yang2019modeling,lee2019set,feng2020point,yan2020pointasnl} to learn the geometric point connections (i.e. attention map) through scalar dot-product. Such a pipeline suffers from high computational cost and struggles to learn large and complex 3D scenes. Point Transformer\cite{zhao2021point, wu2022point}, on the other hand, employs self-attention to local points and uses vector attention  instead to construct the geometric connections between points. This not only requires fewer computations but helps the Point Transformer to learn robust attention weights from high-level geometric connections while encoding local point geometry. The success of Point Transformer shows the importance of point position encoding.

\begin{figure}[!t]
\centering
\includegraphics[width=\columnwidth]{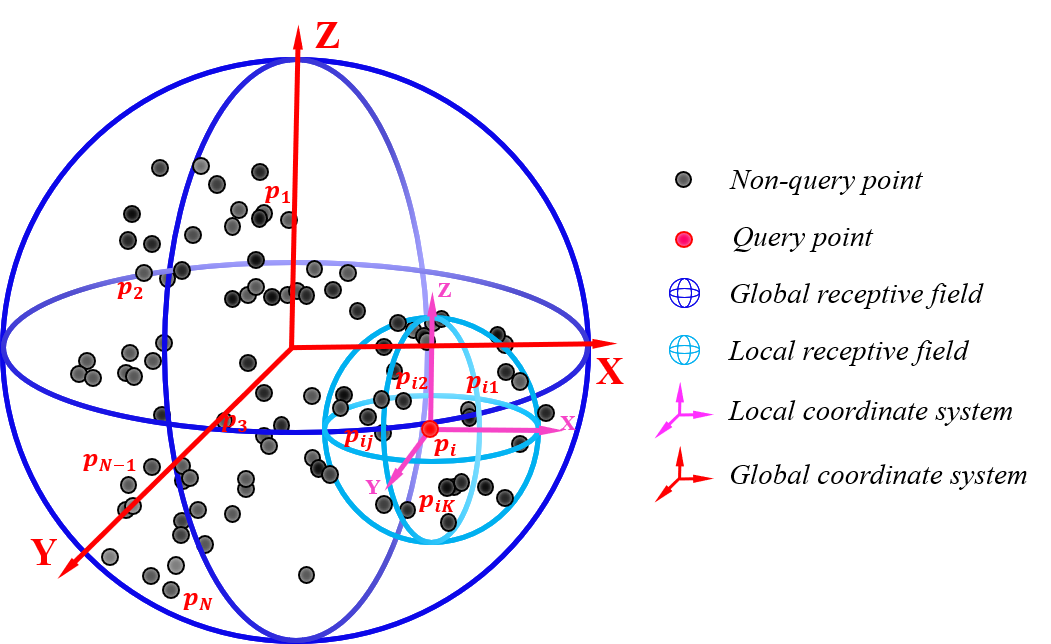}
\caption{Illustration of local and global receptive fields for a query point.}
\label{fig:local_global coordinate}
\end{figure}

\indent Applying the above local aggregation modules can learn the local structure well. However, the local regions divide the global scene into several sub-scenes, breaking the semantic continuity and integrity. Consequently, long-range features (e.g. line features) are not learned effectively because the local features aggregation module cannot exploit the global context. To address this issue, works \cite{feng2020point,ma2020global,yan2020pointasnl} separately extract local features and global context by local and non-local aggregation, and then combine them in a serial and parallel manner. However, these approaches require enormous computations and are still unable to extract the inherent relationship between local features and global context. Motivated by these problems, we propose full point transformer, to simultaneously exploit local features, global context and their correlations.

\section{Method}
\label{section3}
We revisit the three types of local point aggregations i.e., local point-wise MLP, point convolution and point transformer, and then derive a general formulation in Section \ref{section3_subsection1}. Based on the general formulation, we present FPConv in Section \ref{section3_subsection2} and FPTransformer in Section \ref{section3_subsection3} , followed by our down sampling block in Section \ref{section3_subsection4} and finally, in Section \ref{section3_subsection5}, we give our network details.

\subsection{Rethinking Local Point Aggregation}\label{section3_subsection1}
\vspace{-1mm}
Assume we have an unordered point cloud $\{p_{i}\mid i=1,2,\dots,N\}\in\mathbb{R}^{{N\times3}}$  and its corresponding features $\{f_{i}\mid i=1,2,\dots,N\}\in\mathbb{R}^{{N\times C}}$. Here, $p_i$ is a position vector and may also contain additional attributes such as color or surface normal. $N$ and $C$ are the number of points and feature channels respectively. We denote the K neighbors of $p_{i}$ as $\{p_{ij}\mid j=1,2,\dots,K\}\in\mathbb{R}^{{K\times3}}$ and their corresponding features are $\{f_{ij}\mid j=1,2,\dots,K\}\in\mathbb{R}^{{K\times C}}$, as illustrated in Fig. \ref{fig:local_global coordinate}.

\noindent \textbf{Local Point-wise MLP:} The general formulation of local point-wise MLP on a point set $p_{i}$ can be expressed as,
\vspace{-1mm}
\begin{equation}
\mathcal{G}_{i} = \sum_{j=1}^{K}{\rm MLP}(f_{ij}), 
\end{equation}
\vspace{-4mm}

\noindent where MLP is a point-to-point mapping function that maps input features $f_{ij}$ 
to output features $\mathcal{G}_{i}$. 
$\sum$ is a symmetric function e.g. maxpool, sum. The input points have no communication between each other. To establish communication, some works use a function to construct geometric connections between points to enhance point-wise features and then employ an MLP on these enriched point-wise features 

\vspace{-4mm}
\begin{equation}\label{eq2}
\mathcal{G}_{i} =\sum_{j=1}^{K}{\rm MLP}(\mathcal{F}_{1} (p_{ij}, f_{ij})).
\end{equation}
\vspace{-3mm}

\noindent Here, $\mathcal{F}_{1} (.)$ is as a geometric construction function (e.g. hand-crafted geometric descriptor, graph filter, MLP). These functions do not largely increase the network parameters, however, their parameters still need to be optimized, which impacts their ability to generalize across different datasets with varying point densities or shapes.

\noindent \textbf{Local Point Convolution:} Point convolutions introduce the weight function to convolutions such that it learns weights from local point coordinates to dynamically adjust the point-wise features, expressed as

\vspace{-2mm}
\begin{equation}
\mathcal{G}_{i} =\sum_{j=1}^{K}\mathcal{W}(p_{ij})f_{ij}, 
\end{equation}
\vspace{-2mm}

\noindent where  $\mathcal{W}(.)$ is a weight function. Learning weights from the lowest level geometric information (i.e. point coordinates) may not lead to robust convolutional weights. Therefore, some works associate coefficients with weight functions that further adjust the learned weights.
Others propose geometric construction functions that incorporate additional low-level geometric information (e.g. Euclidean distance, position difference, feature difference) to enrich the input to the weight functions, so that they can learn more robust convolutional weights.
These are respectively expressed as,

\vspace{-2mm}
\begin{equation}
\mathcal{G}_{i} = \sum_{j=1}^{K}\mathcal{C}(p_{ij}, f_{ij})\mathcal{W}(p_{ij})f_{ij}, 
\end{equation}
\vspace{-2mm}
\begin{equation}
\mathcal{G}_{i} = \sum_{j=1}^{K}\mathcal{W}(\mathcal{F}_{2}(p_{ij}, f_{ij}))f_{ij}.
\end{equation}
\vspace{-2mm}

\noindent Here $\mathcal{C}(.)$ is the coefficient (also derived from point coordinates or features) of weight function, and $\mathcal{F}_{2}(.)$ is defined as a geometric construction function to construct the rich low level geometric information, such as distance, coordinate difference, and feature difference.

The resolution of point cloud decreases at the deeper encoder layers in a hierarchical network 
leading to the distortion of low level geometric connections derived from sparse points. Learning convolutional weights from such distorted geometric information can lead to non-robust weights.


\noindent \textbf{Local Point Transformer:} Local point transformer solve the above problems and can be expressed as 
\begin{equation}
\begin{aligned}
\mathcal{G}_{i} = \sum_{j=1}^{K}\mathcal{W}(\mathcal{F}_{2}(p_{ij}, f_{ij}) + \delta{(p_{ij})}) (\mathcal{F}_{1} (p_{ij}, f_{ij}) + \delta{(p_{ij})}), 
\end{aligned}
\end{equation}
\noindent where $+$ is the addition operation,  ${\delta{(.)}}$ is the local position encoding that maps the point position in local coordinate system from low-dimension (3D) space to high dimension space. This makes each local point well aware of its position in the local shape. Observing the above formulations of the three classical local feature aggregation methods, we introduce a \textit{general formulation} for local feature aggregation as
\begin{equation}
\begin{aligned}
\mathcal{G}_{i} =\ &\sum_{j=1}^{K}\mathcal{C}(p_{ij},f_{ij})\mathcal{W}(\mathcal{F}_{2}(p_{ij}, f_{ij}) + \delta{(p_{ij})})\\
&(\mathcal{F}_{1}(p_{ij},\!f_{ij}) + \delta{(p_{ij})}). 
\end{aligned}
\end{equation}
Although exiting local aggregation methods get promising performance on local point structures, 
they have two limitations, 
(1) they pay little to no attention to the global geometric structure, and (2) they completely ignore the internal connections between global and local structures. \\ 


\subsection{Full Point Convolution}\label{section3_subsection2}

\textcolor{black}{To exploit local features as well as the global context features along with their correlations, the Full Point Convolution (FPConv) is expressed as}

\begin{equation}\label{eq1}
\mathcal{G}(i) = \sum_{j=1}^{K}\mathcal{W}_c(\mathcal{S}_2(p_{i},p_{ij},\mathcal{S}_1(p_{i},p_{in})))f_{ij},
\end{equation}

\noindent where $S_1$ and  $S_2$ denote the global and local  correlation functions. $\mathcal{W}_c$ is an adaptive weight function that learns the convolutional weights from local and global geometric connections.\\ 
\noindent \textbf{Global Correlation Function:} The goal of global correlation function is to construct the geometric connection between global points and local points, enriching the local points with global geometric information. We define the global correlation function as
\begin{equation}
\mathcal{S}_1(\cdot)=\sum_{n=1}^{N}\mathcal{R}(p_{i},p_{in}), 
\end{equation}
\noindent where $\sum_{n=1}^{N}$ is the aggregation function implemented with summation. $\mathcal{R}(\cdot)$ is the relation between global point $p_{in}$ and each point $p_{i}$, which should be higher when $p_{in}$ is closer to $p_{i}$. Inspired by \cite{thomas2019kpconv}, we propose global linear relation function $\mathcal{C}$
\begin{equation}
\begin{aligned}
\mathcal{R}(\cdot)&=max(0, 1-\frac{||p_{i}-p_{in}||}{\sigma}), 
\end{aligned}
\end{equation}
where $||\cdot||$ is the Euclidena distance between global points and local points. $\sigma$ is the influence coefficient that controls the influence of global points to each point. We set the correlation of global point to center point as $\mathcal{S}_{1i}$ and its corresponding global correlation to neighborhood points as $\mathcal{S}_{1ij}$ .

\noindent \textbf{Local Correlation Function:} Learning convolutional weights highly depends on the local geometric connections. Therefore, we construct sufficient geometric connections using a local connection function. We define the local correlation function as

\begin{equation}
\mathcal{S}_2(\cdot)=p_{ij}+(p_{ij}-p_{i})+||p_{ij}-p_{i}|| + (\mathcal{S}_{1i}-\mathcal{S}_{1ij}), 
\end{equation}
where $p_{i}$ is the center point position, $p_{ij}-p_{i}$ is the position difference, $||p_{j}-p_{i}||$ is the 3D Euclidean distance and $+$ denotes concatenation. Note that there is no learnable parameter in local and global correlation function, hence, they do not bring any computational overhead.\\
\noindent \textbf{Efficient  Weight Function:} The goal of adaptive weight function $\mathcal{W}_c(\cdot)$ is to learn the kernel weights. The output of adaptive weight function are

\begin{figure*}[!t]
\centering
\includegraphics[width= \textwidth]{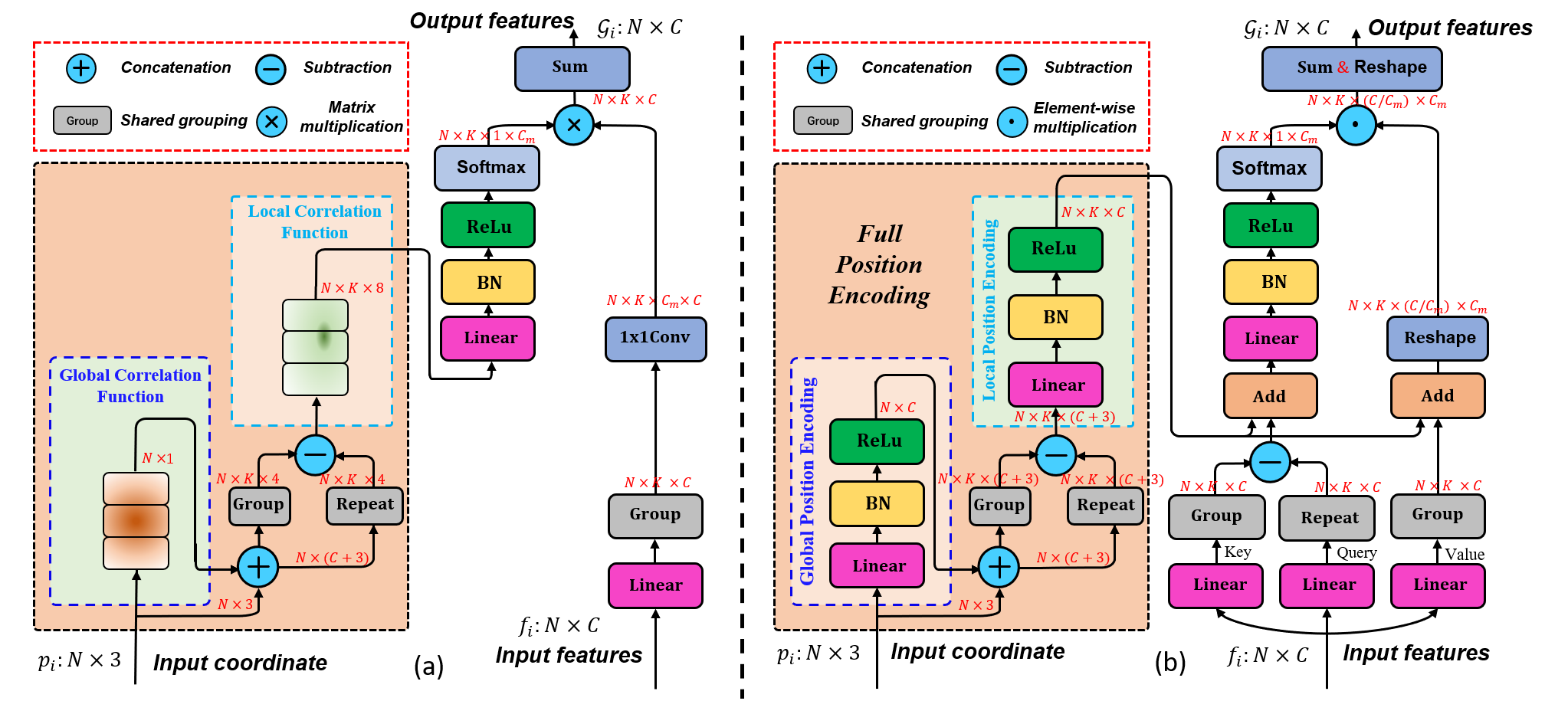}
\vspace{-3mm}
\caption{(a) Proposed Full Point Convolution (FPConv) block which takes the $N\times 3$ input point coordinates and $N\times C$ features from the previous layer to output $N\times C$ features. FPConv incorporates full point correlation (global, local, and global-local correlation) into each point. (b) Proposed Full Point Transformer (FPTransformer) block which takes the $N\times 3$ input point coordinates and $N\times C$ features from the previous layer to output $N\times C$ features. FPTransformer incorporates full position encoding into each point.} 
\label{fig:FPTransformer}
\vspace{-3mm}
\end{figure*}

\begin{equation}
\mathcal{W}_c=\rho_{c}(\phi_c(\mathcal{S}_2(p_{i},p_{ij},\mathcal{S}_1(p_{i},p_{in})))), 
\end{equation}
where {$\phi_c$} is implemented with Multi-layer Perceptrons (MLPs): $\mathbb{R}^{N \times K \times 8}\rightarrow\mathbb{R}^{N \times K\times C\times C} $ and  {$\rho_c$} indicates softmax normalization. Learning a mount of weights from limited geometric connections $\in\mathbb{R}^{K\times 8}$ is inefficient. To this end, we formulate an efficient FPConv based on the following lemma.\\
\noindent\textbf{Lemma:} \emph{FPConv is equivalent to the following formulation: $\mathcal{G}={\rm Max}(\mathcal{T}_{c2}\otimes\rm{Conv_{1\times1}}(\mathcal{T}_{c1},\mathcal{F}_c))$, where $\rm{Conv_{1\times1}}$ is $1\times1$ convolution, $\otimes$ is matrix multiplication and ${\rm Max}$ is maxpooling operation, $\mathcal{T}_{c1}\in\mathbb{R}^{C\times C_{m}\times C}$ is the kernels of $1\times1$ convolution, and  $\mathcal{T}_{c2}\in\mathbb{R}^{K\times C_{m}}$ is the weight matrix learned by adaptive weight function}.\\

\noindent \textbf{Proof:} To better understand the FPConv reformulation, set attention weight matrix $\mathcal{W}_c\in\mathbb{R}^{N \times K \times C}$ as $\mathcal{W}_c(i,j,c)| i=1,...,N,~ j=1,.., K,~ c=1,..,C$ , and set $f_{ij}\in \mathbb{R}^{N \times K\times C}$ as $\mathcal{F}_c(i,j,c)|i=1,...,N,j=1,..,K,c=1,..,C$, where $i$, $j$ and $c$ are the index of the global, neighbor points and input feature channels. According to Eq.\ref{eq1}, FPConv can be expressed as

\begin{equation}\label{eq3}
\mathcal{G}(i)=\sum_{j=1}^{K}\sum_{c=1}^{C}\mathcal{W}_c(i,j,c)\mathcal{F}_c(i,j,c).
\end{equation}
Since the weight function is approximated by MLPs implemented as 1$\times$1 convolutions, the weight matrix generated by weight function can be expressed as
\begin{equation}\label{eq2}
\mathcal{W}_c(i,j,c_{in})=\sum_{c_{m}=1}^{C_{m}}\mathcal{T}_{c2}(i,j,c_{m})\mathcal{T}_{c1}^\mathrm{T}(i,c,c_{m}),
\end{equation}
where $c_{mid}$ and $C_{mid}$ are the index and number of output channels of the middle layer. Substituting Eq.\ref{eq2} into Eq.\ref{eq3}, we get
\begin{equation}\label{eq6}
\begin{aligned}
&\mathcal{G}(i)=\sum_{j=1}^{K}\sum_{c=1}^{C}\mathcal{W}_c(i,j,c)\mathcal{F}_c(i,j,c)\\
&=\sum_{j=1}^{K}\sum_{c=1}^{C}(\sum_{c_{mid}=1}^{C_{m}}\mathcal{T}_{c2}(i,j,c_{m})\mathcal{T}_{c1}^\mathrm{T}(i,c,c_{m}))\mathcal{F}_c(i,j,c)\\
&= \sum_{j=1}^{K}\sum_{c_{m}=1}^{C_{m}}\mathcal{T}_{c2}(i,j,c_{m})\sum_{c=1}^{C}(\mathcal{T}_{c1}^\mathrm{T}(c,c_{mid})\mathcal{F}_c(i,j,c))\\
&={\rm Max}(\mathcal{T}_{c2}\otimes\rm{Conv_{1\times1}}(\mathcal{T}_{c1},\mathcal{F}_c)).
\end{aligned}
\end{equation}

According to the above reformulation, FPConv comprises three operations including one $1\times1$ convolution, one matrix multiplication and one maxpooling. Using this formulation, we divide the kernel weight matrix $\mathcal{W}_c\in\mathbb{R}^{ N\times K\times C\times C}$ into two parts: $1\times1$ convolution weight matrix $\mathcal{T}_{c1}\in\mathbb{R}^{C\times(C_{m}\times C)}$ and weight matrix $\mathcal{T}_{c2}\in\mathbb{R}^{ N\times K\times C_{m}}$, where the $\mathcal{T}_{c1}$ are learned in data driven manner, and $\mathcal{T}_{c2}$ are efficiently learned though the adaptive weight function according to the local and global geometric connections. Here the adaptive weight function can be defined as
\begin{equation}
\mathcal{T}_{2}=\mathcal{\rho}_{ce}(\phi_{ce}(\mathcal{S}_2(p_{i},p_{ij},\mathcal{S}_1(p_{i},p_{in}))), 
\end{equation}
where ${\phi}_{ce}$ is a non-linear function implemented with Multi-layer Perceptrons (MLPs): $\mathbb{R}^{N \times K \times 8}\rightarrow\mathbb{R}^{N \times K\times C_{m}}$. ${\rho}_{ce}$ indicates softmax normalization. The weight function transfers $K \times 8$ dimension geometric connection information to $N \times K \times C_{m}$ dimension weight matrix, where $C_{m}$ is smaller than $C$. With the above reformulation, convolution can learn more robust weights from limited geometric connections. Our reformulted FPConv is shown in Fig.\ref{fig:FPTransformer}(a).



\subsection{Full Point Transformer}\label{section3_subsection3}
We also bring this idea into point Transformer,  we propose FPTransformer that exploits local features as well as the global context and their internal correlations. 
Give a point $(p_i,f_i)$, we use three linear projections to project the point features $f_i$  to the query $q_i$, key $k_i$ and value feature vectors $v_i$, expressed as
\begin{equation}
\begin{aligned}
 {q}_i = {W_q}{f_i}, ~~~~~{k}_i = {W_k}{f_i}, ~~~~~{v}_i = {W_v}{f_i},
\end{aligned}
\end{equation}
where ${W_q}$, ${W_k}$ and ${W_v}$ ($\mathbb{R}^{N \times C}\rightarrow\mathbb{R}^{N \times C}$) are projection functions implemented as linear layers. 
The FPTransformer applied on the point $(p_{i}, f_{i})$ and its corresponding point set $(p_{ij},f_{ij})$ can be formulated as,
\begin{equation}\label{formulation: full point transformer}
\begin{aligned}
 \mathcal{G}_{i} = \sum_{j=1}^{K}\mathcal{W}_a((q_{i}-k_{ij}) + \delta_{full}^{a}) (v_{ij} + \delta_{full}^{a}).
\end{aligned}
\end{equation}
Here, $\sum$ refers to an aggregation function such as summation, 
$\delta_{full}^{a}$ is the full position encoding for FPTransformer, and $\mathcal{W}_a(.)$ is the Attention Function (i.e. Weight Function) that learns the attention map from the geometric connection between key point and query point. 

\noindent \textbf{Full Position Encoding:}
Point features are derived from their coordinates and already contain position information,  however, this information may get diluted in deep aggregation layers. Therefore, 
we add fine-grained position information to features in each aggregation layer. Another limitation of existing position encoding methods is that they tend to map relative position in the local coordinate system from low dimensional (3D)  space to higher dimension. This enhances the awareness of local geometric connections but does not encode the global context. We propose full point position encoding that includes global and local position encoding. 
We define global position encoding on point ${p_i}\in\mathbb{R}^{N\times3}$ as
\begin{equation}
\begin{aligned}
\delta_{global,i}^{a}=[{\phi_{global}^{a}}(p_i),p_i],
\end{aligned}
\end{equation}
the global position encoding function $\phi_{global}^{a}$ is an MLP: $\mathbb{R}^{N \times 3}\rightarrow\mathbb{R}^{N \times C}$. 
$[,]$ is concatenation operation. The output of global position encoding is $\delta_{global,i}^{a}\in\mathbb{R}^{N \times(C+3)}$. we denote the neighborhood point position of $\delta_{global,i}^{a}$ as $\delta_{global,ij}^{a}|j=1,2,...,K$, where $K$ is the number of neighborhood points. The full position encoding on local points $\delta_{global,ij}^{a}\in\mathbb{R}^{N \times K \times(C+3)}$ is defined as
\begin{equation}
\begin{aligned}
\delta_{full}^{a}={\phi_{local}^{a}}(\delta_{global,i}^{a}-\delta_{global,ij}^{a}),
\end{aligned}
\end{equation}
where the local position encoding function $\phi_{local}^{a}$ is an MLP: $\mathbb{R}^{N \times K \times (C+3)}\rightarrow\mathbb{R}^{N \times K \times C}$. 
The output of full position encoding is $\delta_{full}^{a}\in\mathbb{R}^{N \times K \times C}$. With this hierarchical position encoding, each point perceives high dimensional position information in its global and local receptive fields.

\noindent \textbf{Efficient Attention Function:} Attention function plays an important role in our FPTransformer by encoding the geometric connections (i.e. features difference) between query and key points into the attention map. The attention function is formulated as
\vspace{-1mm}
\begin{equation}\label{efficient attention function}
\mathcal{W}_a(\cdot)=\mathcal{\rho}_a(\phi_a((q_{i}-k_{ij}) + \delta_{full}^{a})), 
\end{equation}
\vspace{-1mm}
where {$\phi_{a}$} is an MLP ($\mathbb{R}^{N \times K \times C}\rightarrow\mathbb{R}^{N \times K \times C}$) and {$\rho_{a}$} is softmax normalization to keep the attention weights in the range (0,1). The attention function transfers $N \times K \times C$ geometric connection information to an attention weight matrix $\mathcal{W}_a$ of same dimension. Since, the generation of attention weight matrix $\mathcal{W}_a \in\mathbb{R}^{N \times K \times C} $ requires large memory, we formulate an efficient attention layer as per the following lemma.

\vspace{1mm}
\noindent\textbf{Lemma:} \emph{FPTransformer is equivalent to the following reformulation: $\mathcal{G}_{i}={\rm Sum}(\mathcal{T}_a\odot \mathcal{F}_a)$, where $\mathcal{T}_a\in\mathbb{R}^{N \times K \times 1 \times C_{m}}$ are the attention weights obtained from the attention function, $\mathcal{F}_a\in\mathbb{R}^{N \times K \times C/C_{m}\times C_{m}}$ is the reshaped feature. $\odot$ is element-wise multiplication and ${\rm Sum}$ is summation}.

\noindent \textbf{Proof:} To better understand the FPTransformer reformulation, set attention weight matrix $\mathcal{W}_a\in\mathbb{R}^{N \times K \times C}$ as $\mathcal{W}_a(i,j,c)| i=1,...,N,~ j=1,.., K,~ c=1,..,C$ , and set $(v_{ij} + \delta_{full}^{a})\in \mathbb{R}^{N \times K\times C}$ as $\mathcal{V}(i,j,c)|i=1,...,N,j=1,..,K,c=1,..,C$, where $i$, $j$ and $c$ are the index of the global, neighbor points and input feature channels. As per to Eq.\ref{formulation: full point transformer}, FPTransformer can be expressed as
\vspace{-1mm}
\begin{equation}\label{eq:efficient attention}
\begin{aligned}
&\mathcal{G}_{i}=\sum_{j=1}^{K}\sum_{c=1}^{C}\mathcal{W}_a(i,j,c)\odot\mathcal{V}(i,j,c),\\
&=\sum_{j=1}^{K}\sum_{{c/c_{m}}\times c_{m}=1}^{{C/C_{m}}\times C_{m}}\mathcal{W}_a(i,j,{c/c_{m}}\times c_{m})\odot\mathcal{V}(i,j,{c/c_{m}}\times c_{m}),\\
&=\sum_{j=1}^{K}\sum_{c/c_{m}=1}^{C/C_{m}}\sum_{c_{m}=1}^{C_{m}}\mathcal{W}_a(i,j,c/c_{m},c_{m})\odot\mathcal{V}(i,j,c/c_{m}, c_{m}).
\end{aligned}
\end{equation}

\noindent Here, $C_m$ is the number of middle channels ($C_m < C$), and $c_m$ is the index of middle channel. We set $\{\mathcal{W}_a(i,j,1,c_{m})|i=1,...,N,~~ j=1,...,K,~~ c_{m}=1,...,~~ C_{m}\}\in\mathbb{R}^{C/C_{m}}$ as a vector from attention weight matrix $\mathcal{W}_a\in\mathbb{R}^{N \times K\times C/C_m \times C_m}$, and make these $C/C_{m}$ number of vectors share the same attention weights. Hence, Eq.\ref{eq:efficient attention} can be expressed as
\vspace{-1mm}
\begin{equation}
\begin{aligned}
\mathcal{G}_{i}&=\sum_{j=1}^{K}\sum_{c/c_{m}=1}^{C/C_{m}}\sum_{c_{m}=1}^{C_{m}}\mathcal{W}_a(i,j,c/c_{m},c_{m})\odot\mathcal{V}(i,j,c/c_{m}, c_{m}),\\
&=\sum_{j=1}^{K}\sum_{c/c_{m}=1}^{C/C_{m}}\sum_{c_{m}=1}^{C_{m}}\mathcal{W}_a(i,j,1,c_{m})\odot\mathcal{V}(i,j,c/c_{m}, c_{m}),\\
&={\rm Sum}(\mathcal{T}_a\odot \mathcal{F}_a).
\end{aligned}
\end{equation}
Here, $\mathcal{F}_a\in\mathbb{R}^{ N \times K \times C/C_{m}\times C_{m}}$ is the reshaped feature. $\mathcal{T}_a\in\mathbb{R}^{ N\times K \times 1 \times C_{m}}$ are the attention weights obtained from the efficient attention function defined as
\begin{equation}
\mathcal{W}_a(\cdot)=\rho_{ae}(\phi_{ae}((q_{i}-k_{ij}) + \delta_{full}^{a})), 
\end{equation}
where {${\phi}_{ae}$} is an MLP ($\mathbb{R}^{N \times K \times C}\rightarrow\mathbb{R}^{N \times K \times C_{m}}$) and {$\rho_{ae}$} is softmax normalization. \textcolor{black}{The attention function transfers $N \times K \times C$ dimensional geometric connection information to a $N \times K\times C_{m}$ dimensional attention weight matrix. Similarly, using this reformulation, transformer can learn more robust weights from limited geometric information. Details of $C_m$ are further discussed in Section \ref{section4_subsection4}.}  
Our \textcolor{black}{reformulated} 
FPTransformer is shown in Fig.\ref{fig:FPTransformer}(b).


\subsection{Shape-aware Downsampling (SADS) Block}\label{section3_subsection4}
\begin{figure}[!t]
\centering
\includegraphics[width=0.9\columnwidth]{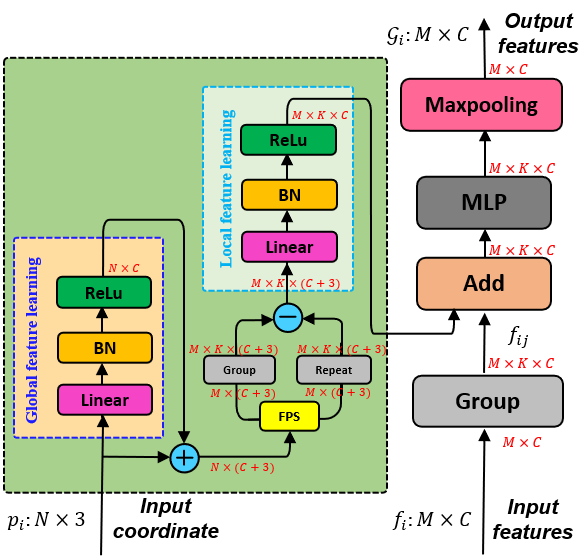}
\caption{Proposed Shape-aware Downsampling (SADS) block. Grouping is done based on the current features to which the full position encoding is added for subsequent downsampling. } 
\label{fig:samplingB}
\vspace{-2mm}
\end{figure}

\vspace{-1mm}
\textcolor{black}{A general downsampling block consists of one sampling and one grouping operation, however, downsampling decreases the resolution of the all points indiscriminately leading the loss of low-level geometric information.} 
\textcolor{black}{To overcome this problem, we propose a shape-aware downsampling (SADS) block with learnable parameters (see Fig.~\ref{fig:samplingB})}.
\textcolor{black}{SADS incorporates a low-level feature learner to maximally preserve the low-level features (i.e. shape details) of the scene}.
Our downsampling block can be expressed as
\begin{equation}
\begin{aligned}
\mathcal{G}_{i} = {\rm max} \left( {\rm MLP}(f_{ij} + H) \right),
\end{aligned}
\end{equation}
where $f_{ij}\in\mathbb{R}^{M\times K \times C}$ are the features from grouping operation, $M$ is the number of points sampled by Farthest Point Sampling (FPS), $H$ are the \textcolor{black}{hierarchical features of points} (see Fig.\ref{fig:samplingB}), and `max' is maxpooling. 
\textcolor{black}{The hierarchical features basically include low-level global and local features. The global features on point ${p_i}\in\mathbb{R}^{N\times3}$ are learned by}
\begin{equation}
\begin{aligned}
f_{global,i}^{d}=[{\phi_{global}^{d}}(p_i),p_i],
\end{aligned}
\end{equation}
where $\phi_{global}^{d}$ is an MLP: $\mathbb{R}^{N \times 3}\rightarrow\mathbb{R}^{N \times C}$. The low-level global features of wise point are $f_{global,i}^{d}\in\mathbb{R}^{N \times(C+3)}$. After sampling and grouping,  we denote the neighborhood point position of sampled point $f_{global,i}^{d}\in\mathbb{R}^{M \times(C+3)}$ as $f_{global,ij}^{d}\in\mathbb{R}^{M \times K \times(C+3)}$. The hierarchical features  $H$ are defined as
\begin{equation}
\begin{aligned}
H={\phi_{local}^{d}}(f_{global,i}^{d}-f_{global,ij}^{d}),
\end{aligned}
\end{equation}
where the local position encoding function is an MLP: $\mathbb{R}^{M \times K \times (C+3)}\rightarrow\mathbb{R}^{M \times K \times C}$. 
\textcolor{black}{This way, simple but effective hierarchical MLPs are integrated into a conventional sampling block.}

\subsection{Network Architecture}\label{section3_subsection5} 

\textcolor{black}{For FPConv, we choose the basic PointNet++\cite{qi2017pointnet++} as the backbone, and replace the MLPs with FPConv to form the new network architectures for four tasks i.e. classification, normal estimation, and semantic segmentation. 
}

\textcolor{black}{For FPTransformer and SADS block,} we use a U-Net like architecture with 5 encoding and decoding layers, and skip connections, \textcolor{black}{covering semantic segmentation, 3D detection, and classification} The first encoding and decoding layers consist of one MLP and FPTransformer. Other encoding layers contain one SADS block and several FPTransformer blocks (details in Section \ref{section4}). 
We set the feature dimensions $C$ as [32,64,128,256,512] and middle feature dimension $C_m$ as [4,8,16,32,64] for the 5 encoding and decoding layers. The ratio of up/down sampling is set as [1,4,4,4,4] for both encoder/decoder.
Decoding layers, besides the first, contain one upsampling block and one FPTransformer. For segmentation, we add an MLP at the end to predict the final point-wise labels. For detection, we use the network as a 3D backbone (excluding the prediction layer). 

\section{Experimental Results}\label{section4}
\textcolor{black}{We evaluate our network on semantic segmentation, 3D detection, shape classification, and normal estimation tasks and perform detailed ablation studies to demonstrate the effectiveness and robustness of the proposed FPConv, FPTransformer and SADS.}

\begin{table*}[t]
\centering
\caption{Semantic segmentation results on S3DIS dataset Area-5. We report the mean class-wise intersection over union (mIoU), mean class-wise accuracy (mAcc) and overall accuracy (OA). 
The best result is in \textbf{Bold}, and second best is \underline{underlined}. }
\label{table: s3dis area5}
\scriptsize
\renewcommand\arraystretch{1.2}
\setlength{\tabcolsep}{0.8mm}{
\begin{tabular}{l|l|c|c|c|c|ccccccccccccc}
\specialrule{1pt}{0pt}{0pt}
\textbf{Year} &\textbf{Method}     & \textbf{mIoU}   & \textbf{mAcc}  & \textbf{OA} &\textbf{Para.}  & \textbf{ceil.} & \textbf{floor} & \textbf{wall}  & \textbf{beam} & \textbf{col.}  & \textbf{wind.} & \textbf{door}  & \textbf{chair} & \textbf{table} & \textbf{book}  & \textbf{sofa}  & \textbf{board} & \textbf{clut.} \\ \hline
2017 CVPR &PointNet\cite{qi2017pointnet} & 41.09  & 48.98 &--  &-- & 88.80 & 97.33 & 69.80 &\textbf{0.05} & 3.92  & 46.26 & 10.76 & 52.61 & 58.93 & 40.28 & 5.85  & 26.38 & 33.22 \\
2018 NIPS &PointCNN\cite{li2018pointcnn}& 57.26    & 63.86 &85.9 &-- & 92.3  & 98.2  & 79.4  & 0.0  & 17.6  & 22.8  & 62.1  & 74.4  & 80.6  & 31.7  & 66.7  & 62.1  & 56.7  \\
2019 ICCV&KPConv\cite{thomas2019kpconv}& 67.1     & 72.8 &-- &--  & 92.8  & 97.3  & 82.4  & 0.0 & 23.9  & 58.0  & 69.0  & \textbf{91.0}  & 81.5  &75.3  & \underline{75.4}  & 66.7  &58.9 \\
2020 PAMI &SPH3D-GCN\cite{lei2020spherical} & 59.5 & 65.9 & -- &--  & 93.3  & 97.1  & 81.1  & 0.0  & 33.2  & 45.8  & 43.8  & 79.7  & 86.9  & 33.2  & 71.5  & 54.1  & 53.7  \\
2020 CVPR &PointANSL\cite{yan2020pointasnl}  &62.6 
&68.5 &87.7 &22.4M &94.3& 98.4 &79.1 & 0.0 &26.7 &55.2 &66.2 &86.8 &83.3 &68.3 &47.6 &56.4 &52.1 \\
2020 CVPR &SegGCN\cite{Lei_2020_CVPR}  &63.6   &70.4 &-- &-- &93.7 &\underline{98.6} &80.6 &0.0 & 28.5 & 42.6 &74.5 & 80.9 &88.7 & 69.0 &71.3 &44.4 &54.3\\
2021 CVPR &PAConv\cite{xu2021paconv} & 66.58 & 73.00   &-- &--  & \underline{94.55} & 98.59 & 82.37 & 0.00 & 26.43 & 57.96 & 59.96 & \underline{89.73} & 80.44 &74.32 & 69.80 & 73.50 & 57.72 \\
2021 CVPR &BAAF-Net\cite{qiu2021semantic} &65.4 &73.1 &88.9 &-- &92.9	&97.9  &82.3	&0.0	&23.1	&\textbf{65.5}	&64.9	&87.5 &78.5	&70.7 &61.4		&68.7	&57.2\\
2022 CVPR &CBL\cite{tang2022contrastive}  &69.4  &75.2 &90.6 &--  &93.9 &98.4 &84.2 &0.0 &37.0 &57.7 &71.9  &81.8 &\underline{91.7}  &75.6 &\textbf{77.8} &69.1 &\textbf{62.9}\\
2022 CVPR &RepSurf-U\cite{ran2022surface} &68.9 &76.0 & 90.2 &1M &--&--&--&--&--&--&--&--&--&--&--&--&--\\
2022 CVPR &Stratified Transformer\cite{lai2022stratified}&\underline{72.0} &\underline{78.1} & \textbf{91.5}   &--&--&--&--&--&--&--&--&--&--&--&--&--\\
2022 ECCV &PointMixer\cite{choe2022pointmixer}  &71.4 &77.4   &-- &-- &94.2 &98.2 &86.0 &0.0 &43.8  &62.1 &\textbf{78.5} &82.2  &90.8 &\underline{79.8}  &73.9 &78.5 &59.4\\
2022 NIPS &PointNeXt\cite{qian2022pointnext} &71.1 &77.2 &91.0 &41.6M &94.2 &98.5 &84.4 &0.0 &37.7 &59.3 &74.0 &91.6 &83.1 &77.2&77.4 &\underline{78.8} &60.6\\
2022 NIPS &Point Transformer V2\cite{wu2022point} &71.6 &77.9 &\underline{91.1} &-- &--&--&--&--&--&--&--&--&--&--&--&--&--\\\hline
2018 NIPS &PointNet++\cite{qi2017pointnet++} &53.4 &62.9 &-- &-- & 89.1 & 98.1 & 73.7 & 0.00 & 3.0 & 58.2 & 21.1 & 67.0 & 78.6 & 44.6 & 60.8 & 56.5 & 43.2 \\
&\textbf{FPConv}(ours) &68.4(\textcolor{blue}{+15})  & 75.0(\textcolor{blue}{+12.1}) &-- &-- & 93.18 & 98.54 & 85.18 & 0.00 & 31.11 & 58.61 & 77.24 & 82.58 & 89.24 & 67.86 & 73.45 & 74.95 &56.89 \\\hline
2021 ICCV &Point Transformer\cite{zhao2021point}&70.4   &76.5 &90.8  &7.8M &94.0 &98.5 &\underline{86.3} &0.0 & 38.0 &63.4 &74.3  &82.4 &89.1  &\textbf{80.2} &74.3 &76.0 &59.3\\ 
&\textbf{FPTransformer}(ours)  &\textbf{72.2}(\textcolor{blue}{+1.8}) &\textbf{78.5}(\textcolor{blue}{+2.0}) &\textbf{91.5}(\textcolor{blue}{+0.7}) &10.9M &\textbf{94.6} & \textbf{98.7} & \textbf{87.2} & 0.00 & \textbf{44.5} & \underline{64.8} & \underline{77.0} & 82.9 & \textbf{92.3} &  79.3 &74.6 & \textbf{81.2} &\underline{61.8}\\
 \specialrule{1pt}{0pt}{0pt}
\end{tabular}}
\end{table*}

\begin{table*}[t]
\centering
\caption{Semantic segmentation results on S3DIS with 6-fold cross validation. We report the mean class-wise IoU (mIoU), mean class-wise accuracy (mAcc) and overall accuracy (OA).
Best result is in \textbf{Bold}, and second best is \underline{underlined}.} 
\label{table: s3dis 6-fold}
\scriptsize
\renewcommand\arraystretch{1.2}
\setlength{\tabcolsep}{1.1mm}{
\begin{tabular}{l|l|c|c|c|ccccccccccccc}
\specialrule{1pt}{0pt}{0pt}
\textbf{Year} &\textbf{Method}    & \textbf{mIoU}   & \textbf{mAcc}   & \textbf{OA}  & \textbf{ceil.} & \textbf{floor} & \textbf{wall}  & \textbf{beam} & \textbf{col.}  & \textbf{wind.} & \textbf{door}  & \textbf{chair} & \textbf{table} & \textbf{book}  & \textbf{sofa}  & \textbf{board} & \textbf{clut.} \\ \hline
2017 CVPR &PointNet\cite{qi2017pointnet} &47.6  &66.2 &78.6  &88.0 &88.7 &69.3 &42.4 &23.1 &47.5 &51.6 &42.0 &54.1 &38.2 &9.6 &29.4 &35.2 \\
2018 NIPS &PointCNN\cite{li2018pointcnn}&65.4 &75.6 &88.1 &\underline{94.8} &97.3 &75.8 &63.3 &51.7 &58.4 &57.2 &69.1 &71.6  &61.2 &39.1  &52.2 &58.6  \\
2019 CVPR &PointWeb\cite{zhao2019pointweb} &66.7   &76.2 &87.3  &93.5 &94.2 &80.8 &52.4 &41.3 &64.9 &68.1 &67.1 &71.4  &62.7 &50.3  &62.2 &58.5\\
2019 ICCV &KPConv\cite{thomas2019kpconv}&70.6 &79.1 &--  &93.6 &92.4 &83.1 &\underline{63.9} &54.3 &66.1 &76.6 &64.0 &57.8  &\underline{74.9} &69.3 &61.3 &60.3\\
2020 PAMI &SPH3D-GCN\cite{lei2020spherical}&68.9 &77.9 &88.6  &93.3 &96.2 &81.9 &58.6 &55.9 &55.9 &71.7 &\underline{82.4} &72.1  &64.5 &48.5  &54.8 &60.4\\
2020 CVPR &PointANSL\cite{yan2020pointasnl}  &68.7 
&79.0 &88.8  & 95.3 &97.9 &81.9 &47.0 &48.0 &67.3 &70.5 &77.8 &71.3 &60.4 &50.7 &63.0 &62.8 \\
2020 CVPR &RandLA-Net\cite{hu2020randla} &70.0 &82.0 &88.0  &93.1 &96.1 &80.6 &62.4 &48.0 &64.4 &69.4 &76.4 &69.4  &64.2 &60.0  &65.9 &60.1\\
2021 CVPR &PAConv\cite{xu2021paconv}&69.31   &78.65 &--  &94.30 &93.46 &82.80 &56.88 &45.74 &65.21 &74.90 &59.74 &74.60 &67.41 &61.78 &65.79 &58.36\\
2021 CVPR &SCF-Net\cite{fan2021scf}  &71.6 &82.7 &88.4  &93.3 &96.4 &80.9 &\textbf{64.9} &47.4 &64.5 &70.1  &81.6 &71.4 &64.4 &67.2  &\underline{67.5} &60.9\\
2021 CVPR &BAAF-Net\cite{qiu2021semantic} &72.2 &\underline{83.1} &88.9  &93.3 &96.8 &81.6 &61.9 &49.5 &65.4 &73.3  &\textbf{83.7} &72.0 &64.3 &67.5  &67.0 &62.4\\
2022 NIPS &PointNeXt\cite{qian2022pointnext} &\underline{74.9} &83.0 &90.3  &--&--&--&--&--&--&--&--&--&--&--&--&--\\
2022 CVPR &RepSurf-U\cite{ran2022surface} &74.3 &82.6 &\underline{90.8}  &--&--&--&--&--&--&--&--&--&--&--&--&--\\

2022 CVPR &CBL\cite{tang2022contrastive} &73.1 &79.4  &89.6  &94.1 &94.2 &\underline{85.5} &50.4 &\underline{58.8} &\textbf{70.3} &\underline{78.3} &75.0 &75.7  &74.0 &\textbf{71.8}  &60.0 &62.4 \\ \hline

2021 ICCV &Point Transformer \cite{zhao2021point}&73.5  &81.9 &90.2  &94.3 &\underline{97.5} &84.7 &55.6 &58.1 &66.1 &78.2 &74.1 &\underline{77.6} &71.2 &67.3 &65.7 &\underline{64.8} \\
&\textbf{FPTransformer}(ours) &\textbf{76.0}(\textcolor{blue}{+2.5})  & \textbf{84.0}(\textcolor{blue}{+2.1}) &\textbf{91.0}(\textcolor{blue}{+0.8}) & \textbf{95.0} & \textbf{97.6} & \textbf{86.0} & 59.7 &\textbf{60.2} & \underline{68.3} & \textbf{78.6} & 75.2 & \textbf{84.3} &\textbf{75.2} &\underline{69.6} & \textbf{71.5} &\textbf{67.0} 
\\\specialrule{1pt}{0pt}{0pt}
\end{tabular}}
\end{table*}

\begin{figure*}[!t]
\centering
\includegraphics[width= \textwidth]{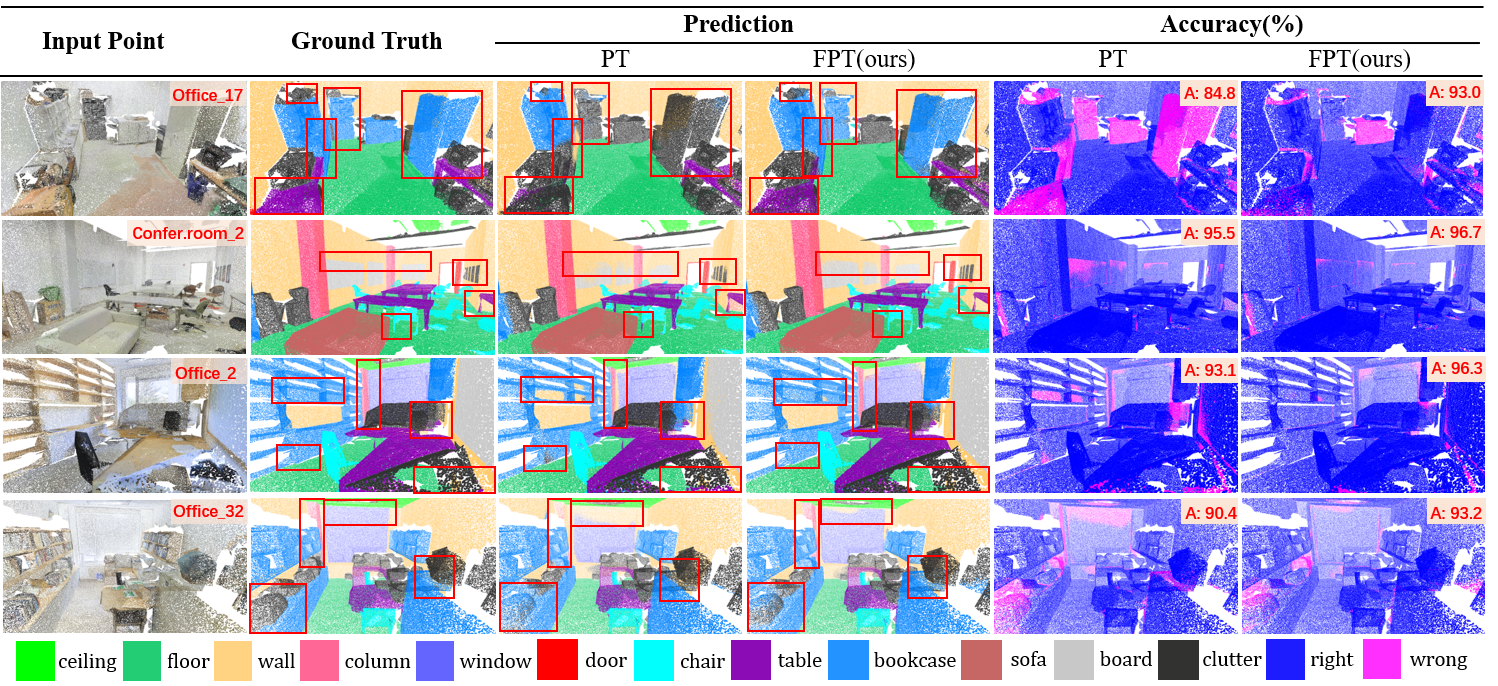}
\vspace{-6mm}
\caption{Visualization of semantic segmentation results on S3DIS Area-5. 
The \textcolor{red}{red} boxes highlight areas in the scenes where our proposed FPTransformer performs particularly better than the Point Transformer (PT).} 
\label{fig:s3disVis}
\end{figure*}

\subsection{Semantic Segmentation}
\noindent \textbf{Datasets:} 
We evaluate our network on two large-scale indoor scene datasest, S3DIS and ScanNet. S3DIS \cite{armeni20163d} contains colored point clouds annotated point-wise with 13 classes. It covers 271 rooms from 6 large-scale indoor scenes (total of 6020 square meters).  We conduct 6-fold cross validation on S3DIS and, in line with other works, conduct more extensive comparisons on Area 5 as the test set, which is not in the same building as the other areas. ScanNet\cite{dai2017scannet} contains colored point clouds of indoor scenes with point-wise semantic labels of 20 object categories. It is split into 1201 scenes for training and 312 for validation.
%
%
\begin{table}[t]
\centering
\caption{Semantic segmentation results (mIoU) on ScanNet Validation.}
\label{table: scannet sementic}
\scriptsize
\renewcommand\arraystretch{1.2}
\setlength{\tabcolsep}{1mm}{
\begin{tabular}{l|l|c|c|c}
\specialrule{1pt}{0pt}{0pt}
\textbf{Year} &\textbf{Method}    &\textbf{Input} & \textbf{\#Parameter} & \textbf{mIoU}(\%)    \\ \hline
2018 NIPS &PointNet++\cite{qi2017pointnet++} &point &-- & 53.5  \\
2018 CVPR &SparseConvNet\cite{graham20183d}  &voxel &-- & 69.3   \\
2019 CVPR &PointConv\cite{wu2019pointconv}  &point &-- &61.0    \\
2020 CVPR &PointANSL\cite{yan2020pointasnl}  &point&-- &63.5  \\
2019 ICCV &MVPNet\cite{jaritz2019multi}  &point &-- & 66.4  \\
2019 ICCV &KPConv\cite{thomas2019kpconv}  &point &14.9M &69.2 \\
2019 3DV &JointPointBased\cite{chiang2019unified} &point &-- &69.2 \\
2022 CVPR &RepSurf-U\cite{ran2022surface} & point &-- &70.0 \\
2022 CVPR &Stratified Transformer\cite{lai2022stratified} &point & 18.8M &\textbf{74.3} \\ 
2021 CVPR &BPNet\cite{hu2021bidirectional} &point+image&-- &72.5 \\
2019 CVPR &MinkowshiNet\cite{choy20194d}  &voxel &37.9M &72.2\\
2022 CVPR &FastPointTransformer\cite{park2022fast} &voxel &37.9M &72.0 \\
\hline
2021 CVPR &PointTransformer\cite{zhao2021point} &point &10.7M &70.6  \\
&\textbf{FPTransformer} (ours) &point &12.2M &\underline{73.9}(\textcolor{blue}{+3.3})\\
\specialrule{1pt}{0pt}{0pt}
\end{tabular}}
\end{table}

\noindent \textbf{Network Configuration:} 
For semantic segmentation on S3DIS, we set the FPTransformer block in the 5 encoding layers depths [1,2,2,6,2]. We set the voxel size as 4cm and the maximum number of voxels to 80,000. We adopt the SGD optimizer and weight decay as 0.9 and 0.0001. The base learning rate is set as 0.5 and learning rate is scheduled by the MultiStepLR every 30 epochs. We train and test the model with batch size 16 on 4 GPUs and batch size 4 on a single GPU, respectively. We adopt scaling, 
contrast, translation, jitter and chromatic translation 
to augment training data.  
On ScanNet, we set the FPTransformer block in the 5 encoding layer depths as [1,3,9,3,3]. We set the voxel size as 2cm and the maximum number of voxels as 120,000. The base learning rate is set as 0.1. We use rotation, flip, scaling, and jitter
to augment training data. 

\noindent \textbf{Results:} We compare our method with the recent state-of-the-art on three metrics i.e. mean class-wise intersection over union (mIoU), mean overall accuracy (mAcc) and network parameters (Para.). \textcolor{black} {Table \ref{table: s3dis area5} provides detailed results on S3DIS Area-5. Our network equipped with FPTransformer and shape-aware downsampling (SADS) achieves the best performance 72.2\%, 78.5\% and 91.5\% in terms of mIoU, mAcc and OA, respectively. It also achieves competitive results (top 2) on 9 out of the 13 categories including \textit{ceiling, floor, wall, column, window, door, table, board and clutter}. Moreover, compared to the classical vector attention based method (i.e. Point Transformer\cite{zhao2021point}), the performance of our method exceeds it by a large margin on some long-range shape classes  such as \textit{column, door, table, board}. Compared to the latest state-of-the-art network PointNeXt, our method not only outperforms it on mIoU and mAcc, but also has 74\% less network parameters. Compared to the classical vector attention method (e.g. Point Transformer and Point Transformer V2), our FPTransformer gets significant improvements. Beside, our FPConv achieves the best performance among classical Convolution-based method such as KPConv and PAConv, and exceed the baseline PointNet++ 15\% in the term of mIoU.}

\textcolor{black}{We report results in Table \ref{table: s3dis 6-fold} for the 6-fold validation setting on S3DIS dataset. Our method again performs the best, achieving state-of-the-art results on all three metrics i.e. 76.0\% mIoU, 84.0\% mAcc and 91.0\% OA. Notably, our method achieves the best performance on 9 out of the 13 categories and achieves the second best performance on another 2 categories.} 
Fig.~\ref{fig:s3disVis} shows visualizations of our results on S3DIS area 5 in comparison to Point Transformer. We can see that our method is more robust to long-range shapes such as book case and board. 

Table \ref{table: scannet sementic} shows semantic segmentation results on ScanNet validation set and compares the proposed FPTransformer to existing state-of-the-art. Compared to Point Transformer (which also uses vector attention), our method gets significant improvement of +3.3 in mIoU  (Table \ref{table: scannet sementic} last row). Our method even outperforms the multi-modal BPNet\cite{hu2021bidirectional}. Although our method performs slightly lower than Stratified Transformer, our network has about 35\% fewer parameters. Compared to the popular voxel based methods FastPointTransformer\cite{park2022fast} and MinkowshiNet\cite{choy20194d}, our method has higher mIoU and less than half the network parameters. In this experiment, our network used 2cm voxel size. More comparisons for varying voxel size are provided in the ablation study.

\subsection{3D Object Detection}
We conduct experiments for 3D object detection to show the generalization ability of FPTransformer.

\noindent \textbf{Datasets:} 
We conduct experiments on two popular datasets:  ScanNetV2\cite{dai2017scannet}  and KITTI\cite{geiger2013vision}. 
ScanNetV2 consists of 1513 indoor scenes and 18 object classes. On this dataset, we adopt mean Average Precision (mAP) metric under the threshold of 0.25 (mAP@0.25) and 0.5 (mAP@0.5) without considering the bounding box orientations. 
KITTI is a large scale outdoor dataset captured with LiDAR sensor. It has 7518 test and 7481 training samples (further divided into train-validation splits). We calculate mAP for easy, moderate, and hard cases, at 11 and 40 recall positions following the official KITTI protocol.

\begin{table}[t]
\centering
\caption{Detection results on ScanNetV2 validation set from the benchmark website. We report mAP at thresholds of 0.25 and 0.5 IoU. * denotes model reproduced by \cite{contributors2020mmdetection3d}.}
\label{table: scannet detection}
\scriptsize
\renewcommand\arraystretch{1.2}
\setlength{\tabcolsep}{0.7mm}{
\begin{tabular}{l|l|l|c|c}
\specialrule{1pt}{0pt}{0pt}
\textbf{Year} &\textbf{Method}    &\textbf{3D Backbone} & \textbf{mAP0.25} & \textbf{mAP0.5}    \\ \hline
2020 ECCV &H3DNet\cite{zhang2020h3dnet}   & 4$\times$PointNet++ &67.2 &48.1  \\
2021 ICCV &3DETR\cite{misra2021end}   &Transformer&65.0 &47.0  \\
2021 CVPR &BRNet\cite{cheng2021back}  &4$\times$PointNet++  &66.1 &50.9  \\
2022 CVPR &TokenFusion\cite{wang2022multimodal}   &2$\times$PointNet++  &70.8 &54.2  \\
2022 CVPR & RBGNet\cite{wang2022rbgnet}  &PointNet++ &70.6 &\textbf{55.2}  \\
2022 CVPR &RepSurf-U\cite{ran2022surface}   &PointNet++ &71.2 &54.8 \\\hline
2019 CVPR &VoteNet\cite{qi2019deep}  &PointNet++ &58.6 & 33.5  \\
 &VoteNet (ours) &\textbf{FPTransformer} &62.3(\textcolor{blue}{+3.7}) & 38.6(\textcolor{blue}{+5.5})  \\
2019 CVPR &VoteNet*\cite{qi2019deep}  &PointNet++ &62.9 &39.9   \\
 &VoteNet* (ours)  &\textbf{FPTransformer} &65.2(\textcolor{blue}{+2.3}) &45.1(\textcolor{blue}{+5.2}) \\\hline
 2021 CVPR &Group-Free-3D(6,256)\cite{liu2021group}  &PointNet++ &67.3 &48.9 \\
 &Group-Free-3D(6,256) (ours)  &\textbf{FPTransformer} &69.2(\textcolor{blue}{+1.9}) &50.7(\textcolor{blue}{+1.8}) \\
 2021 CVPR &Group-Free-3D(12,512)\cite{liu2021group}  &2$\times$ PointNet++ &69.1 &52.8 \\
 &Group-Free-3D(12,512)(ours)  &\textbf{FPTransformer} &\textbf{71.5}(\textcolor{blue}{+2.4}) &54.3(\textcolor{blue}{+1.5}) \\\specialrule{1pt}{0pt}{0pt}
\end{tabular}}
\vspace{-4mm}
\end{table}

\begin{table*}[h]
\caption{3D object detection results on the KITTI validation set. We report mean Average Precision (mAP)  at 11 and 40 recall positions, similar to prior works \cite{shi2019pointrcnn, shi2020pv}.}
\label{table: kitti detection}
\scriptsize
\centering
\renewcommand\arraystretch{1.1}
\setlength{\tabcolsep}{0.8mm}{
\begin{tabular}{l|l|c|c|c|c|c|c|c|c|c|c}
\specialrule{1pt}{3pt}{0pt}
  \multicolumn{1}{c|}{\multirow{2}{*}{\textbf{Methods}}} &\multicolumn{1}{c|}{\multirow{2}{*}{\textbf{3D Backbone}}} & \multicolumn{3}{c|}{\textbf{Car}(\%)} 
  & \multicolumn{3}{c|}{\textbf{Pedestrian}(\%)} 
  & \multicolumn{3}{c|}{\textbf{Cyclist}(\%)}&\multicolumn{1}{c}{\multirow{2}{*}{\textbf{Recall}}}\\ \cline{3-11}
 &  & Easy & Moderate & Hard  & Easy & Moderate & Hard & Easy & Moderate & Hard \\ \hline
 
PointRCNN\cite{shi2019pointrcnn} &PN++(MSG) &88.72 &78.55 &77.72 &62.64 &55.06 &50.83 &\textbf{85.68} &\textbf{71.22} &\textbf{65.63} &\multirow{6}{*}{R@11}  \\ 
PointRCNN\cite{shi2019pointrcnn} &PT &88.89 &78.62 &77.94 &65.53 &58.24 &55.76 &83.95 &69.45 &64.97  \\ 
PointRCNN(ours) &\textbf{FPTrans.} &\textbf{89.05}(\textcolor{blue}{+0.33}) &\textbf{78.89}(\textcolor{blue}{+0.44}) &\textbf{78.44}(\textcolor{blue}{+0.72}) &\textbf{68.78}(\textcolor{blue}{+6.14}) &\textbf{61.45}(\textcolor{blue}{+6.39}) &\textbf{56.57}(\textcolor{blue}{+5.74}) &84.10(\textcolor{blue}{-1.58}) &70.15(\textcolor{blue}{-1.07}) &65.55(\textcolor{blue}{-0.08}) \\\cline{1-11}

PV-RCNN\cite{shi2020pv} &Spconv + MLP &89.23 &83.23 &78.81 &67.29 &61.01 &56.40 &87.16 &73.45 &68.98 \\
PV-RCNN\cite{shi2020pv} &Spconv + PT &89.10 &82.98 &77.76 &67.75 &61.20 &56.43 &86.64 &72.54 &69.11 \\
PV-RCNN (ours) &Spconv + \textbf{FPTrans.} &\textbf{89.43}(\textcolor{blue}{+0.2}) &\textbf{83.60}(\textcolor{blue}{+0.37}) &\textbf{78.84}(\textcolor{blue}{+0.03}) &\textbf{68.21}(\textcolor{blue}{+0.92}) &\textbf{61.45}(\textcolor{blue}{+0.44}) &\textbf{56.62}(\textcolor{blue}{+0.22}) &86.59(\textcolor{blue}{-0.57}) &72.32(\textcolor{blue}{-1.13}) & \textbf{69.27}(\textcolor{blue}{+0.29}) \\\cline{1-12}

PointRCNN\cite{shi2019pointrcnn} &PN++(MSG) &91.28 &80.47 &78.02 &62.80 &55.41 &48.83 &90.58 &70.98 &66.65  &\multirow{6}{*}{R@40} \\   
PointRCNN\cite{shi2019pointrcnn} &PT &91.10 &80.23 &77.96 &65.74 &58.65 &53.45 &90.68 &71.04 &66.84  \\  
PointRCNN (ours) &\textbf{FPTrans.} &\textbf{91.30}(\textcolor{blue}{+0.02}) &\textbf{81.87}(\textcolor{blue}{+1.40}) &\textbf{80.14}(\textcolor{blue}{+2.12}) &\textbf{68.55}(\textcolor{blue}{+5.75}) &\textbf{61.98}(\textcolor{blue}{+6.57}) &\textbf{56.17}(\textcolor{blue}{+7.34}) &\textbf{91.76}(\textcolor{blue}{+1.18}) &\textbf{71.45}(\textcolor{blue}{+0.57}) &\textbf{67.24}(\textcolor{blue}{+0.59}) \\\cline{1-11}

PV-RCNN\cite{shi2020pv} &Spconv + MLP &92.00 &84.38 &82.42 &68.19 &60.55 &55.55 &90.39 &70.38 &65.97 \\ 
PV-RCNN\cite{shi2020pv} &Spconv + PT &92.04 &84.45 &82.43 &68.23 &60.76 &56.23 &91.45 &71.23 &67.56 \\ 

PV-RCNN (ours) &Spconv + \textbf{FPTrans.}  & \textbf{92.15}(\textcolor{blue}{+0.15}) &\textbf{84.74}(\textcolor{blue}{+0.36}) &\textbf{82.62}(\textcolor{blue}{+0.20}) &\textbf{68.24}(\textcolor{blue}{+0.05}) &\textbf{61.10}(\textcolor{blue}{+0.55}) &\textbf{57.32}(\textcolor{blue}{+1.77}) &\textbf{92.71}(\textcolor{blue}{+2.32}) &\textbf{72.50}(\textcolor{blue}{+2.12}) & \textbf{69.26}(\textcolor{blue}{+3.29})\\\specialrule{1pt}{0pt}{0pt}    
\end{tabular}
}
\end{table*}

\begin{figure*}[!t]
\vspace{-3mm}
\centering
\includegraphics[width= 0.8\textwidth]{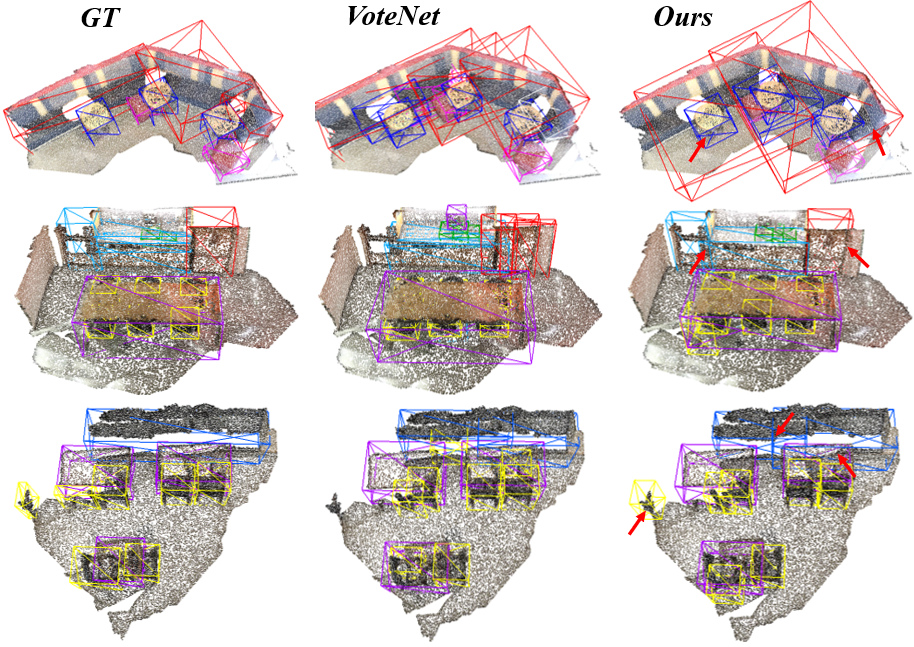}
\caption{Visualization of detection results on ScanNetV2. The \textcolor{red}{red} arrows indicate objects in the scenes where the VoteNet equipped with FPTransformer performs better than  original VoteNet.} 
\label{fig:scannetVis}
\vspace{-3mm}
\end{figure*}

\noindent \textbf{Network Configuration:} 
For ScanNetV2, we select two detection architectures: VoteNet\cite{qi2019deep} and Group-free-3D\cite{liu2021group}. For KITTI, we choose PointRCNN\cite{shi2019pointrcnn} and PVRCNN\cite{shi2020pv}. 
VoteNet, Group-Free-3D and PointRCNN have a similar network architecture. They use a 3D backbone to extract point features for subsequent object bounding box prediction. 
We replace their 3D backbones with our network. 
PVRCNN integrates voxel and point features, where the voxel features are  learned by 3D sparse convolution using multiple encoding layers and summarized into a small set of key points via the voxel set abstraction module. 
We replace the abstraction module with our FPTransformer. 
During training, we keep the same parameters of the original PointRCNN and PVRCNN networks. However, for VoteNet, we change the number of input points to 40,960 and epochs to 260. For Group-Free-3D, we only adopt the point coordinates as input features.

\noindent \textbf{Results:} As shown in Table \ref{table: scannet detection}, when FPTransformer is plugged into the VoteNet and Group-Free-3D networks, the detection performance improves significantly on ScanNetV2. Specifically, VoteNet official model with FPTransformer obtains 3.7\% mAP@0.25 and 5.5\% mAP@0.5 improvements. Similarly, 
the VoteNet reproduced by \cite{contributors2020mmdetection3d} with FPTransformer obtains 2.3\% mAP@0.25 and 5.2\% mAP@0.5 improvements on the MMDetection3D platform. As shown in Figure \ref{fig:scannetVis}, FPtransformer helps VoteNet to reduce false positive detection.
Our FPTransformer backbone also improves the accuracy of both versions of the the Group-Free-3D detection network. Specifically, the Group-Free-3D (12 decoder layers, 512 object candidates) with our FPTransformer backbone gains 2.4\% mAP@0.25 and 1.5\% mAP@0.5 improvement, outperforming all prior models equipped with transformer modules such as 3DETR\cite{misra2021end} and TokenFusion\cite{wang2022multimodal}, on the mAP@0.25 metric.

Table \ref{table: kitti detection} shows results on the KITTI dataset. Our method consistently improves the performance of both detectors for Recall@40 and improves 13 out of 18 cases for Recall@11. PointRCNN with FPTransformer obtains remarkable improvements of 5.75\%, 6.57\%, 7.34\% (Recall@40) on the easy, moderate, hard cases of the `Pedestrian' class. 
PVRCNN with FPTransformer obtains significant improvements of 1.77\%, 3.29\% for the hard cases of `Pedestrian', `Cyclist'.

\subsection{Classification}\label{section4_subsection3}
\noindent \textbf{Synthetic Data:} We firstly evaluate FPConv and FPTransformer on ModelNet40\cite{wu20153d} which comprises 12,311 CAD models from 40 categories. The data is divided into 9,843 training and 2,468 test models. We uniformly sample 1024 points from each models and only use their $(x,y,z)$ coordinates as input. The training data is augmented by randomly translating in the range $[-0.2, 0.2]$, scaling in the range $[0.67, 1.5]$.

\noindent \textbf{Real-world Data:} 
We also evaluate our network on the real-world ScanObjectNN dataset\cite{uy2019revisiting} which includes 15,000 objects categorized into 15 classes. Unlike the synthetic ModelNet40 objects, these objects are with occlusion, background noise, deformed geometric shapes and non-uniform surface density providing a more challenging scenario. We conduct experiments on its hardest perturbed variant (i.e. PB$\_$T50$\_$RS variant). We uniformly sample 1024 points from each object and only use their $(x,y,z)$ coordinates as input. The training data is augmented similar to ModeleNet40.

\noindent \textbf{Network Configuration:} We use the same network configuration for ModelNet40 and ScanOjectNN. 
During training, we use SGD optimizer with 0.9 momentum and 0.1 initial learning rate to train our model for 350 epochs with a batch size of 32. We adopt cosine annealing to dynamically adjust the learning rate when it drops to 0.001 and use a dropout ratio of 0.4. 

\begin{table}[htbp]
\centering
\caption{Classification results on ModelNet40 and ScanObjectNN dataset. Our network achieves the best overall accuracy. ‘xyz’ and ‘n’ represent coordinates and normal vector. ‘K' stands for one thousand. ‘PN++’ stands for PointNet++ and ‘*’ denotes methods evaluated with voting strategy \cite{liu2019relation}.}
\label{table1}\label{tab1}
\scriptsize
\renewcommand\arraystretch{1.1}
\setlength{\tabcolsep}{2.2mm}{
\begin{tabular}{l|c|c|c|c}
\specialrule{1pt}{0pt}{0pt}
\multirow{2}{*}{\textbf{Method}}  &\multirow{2}{*}{\textbf{Input}}&\multirow{2}{*}{\textbf{\#Points}}  
& \multicolumn{2}{c}{\textbf{OA}(\%)} \\\cline{4-5}  & & & ModelNet40          & ScanObjectNN          \\\hline
PointWeb\cite{zhao2019pointweb}       & xyz, n & 1K       & 92.3 &-\\
PointConv\cite{wu2019pointconv}     & xyz, n & 1K       & 92.5 &-\\
SpiderCNN\cite{xu2018spidercnn}        & xyz, n & 5K       & 92.4 &-\\
KPConv\cite{thomas2019kpconv}        & xyz      & 7K       & 92.9 &-\\
PointASNL\cite{yan2020pointasnl}      & xyz, n & 1K       & 93.2 &-\\ 
PRANet\cite{cheng2021net}      & xyz      & 2K      & 93.7 &\underline{82.1}\\\hline
RS-CNN\cite{liu2019relation}      & xyz      & 1K       & 92.9 &-  \\
RS-CNN*\cite{liu2019relation}      & xyz      & 1K       & 93.6 &-  \\
3DmFV\cite{ben20183dmfv}        & xyz      & 1K       & 91.4 &63.0\\
PointNet\cite{qi2017pointnet}          & xyz      & 1K       & 89.2 &68.2\\
PointNet++\cite{qi2017pointnet++}     & xyz      & 1K       & 90.7 &77.9\\
DGCNN\cite{wang2019dynamic}             & xyz      & 1K       & 92.9 &78.2\\
PointCNN\cite{li2018pointcnn}  & xyz      & 1K       & 92.2 &78.5\\
BGA-DGCN\cite{uy2019revisiting}        & xyz      & 1K       & - &79.9\\ 
BGA-PN++\cite{uy2019revisiting}       & xyz      & 1K       & - &80.2\\ 
DRNet\cite{qiu2021dense}      & xyz      & 1K       & 93.1 &80.3\\ 
GBNet\cite{qiu2021geometric}      & xyz      & 1K       & 93.8 &80.5\\ 
PointASNL\cite{yan2020pointasnl}      & xyz & 1K       & 92.9 &-\\
PRANet\cite{cheng2021net}      & xyz      & 1K      & 93.2 &81.0\\ \hline
PointNet++\cite{qi2017pointnet++} & xyz      & 1K       & 90.7 &77.9\\
\textbf{FPConv}(ours)       & xyz      & 1K       & 93.6 (\textcolor{blue}{+2.9}) &84.6 (\textcolor{blue}{+6.7})\\ 
\textbf{FPConv*}(ours)      & xyz      & 1K       & \underline{93.9} (\textcolor{blue}{+3.2}) &\underline{85.6}
(\textcolor{blue}{+7.7})\\  \hline
PointTransformer\cite{zhao2021point}       & xyz      & 1K       &93.7  &--\\ 
\textbf{FPTransformer}(ours)       & xyz      & 1K       &\textbf{94.1}(\textcolor{blue}{+0.4})  &\textbf{86.0}\\\specialrule{1pt}{0pt}{0pt}
\end {tabular}}
\vspace{-4mm}
\end{table}

\noindent \textbf{Results:} We compare our method with representative state-of-the art methods in Table \ref{tab1} using the overall accuracy (OA) metric. For better comparison, we also show input data type and number of input points for each method. \textcolor{black}{The FPConv network} achieves competitive OA of 93.9\% on ModelNet40 and 85.6\% on ScanObjectNN, giving significant improvements of 3.2\% and 7.7\% over the backbone PointNet++\cite{qi2017pointnet++}.
On ModelNet40, our network outperforms the classical local point convolution KPConv \cite{thomas2019kpconv}, 
by 1\% even though KPConv uses 7,000 input points and our network only uses 1,024 points. Moreover, our network outperforms RS-CNN when voting strategy is not used. 

\textcolor{black}{The FPTransformer network achieves the stat-of-the-art OA of 94.1\% on ModelNet40 and 86.0\% on ScanObjectNN. On ModelNet40,} our network outperforms vector attention based method (e.g. Point Transformer)
by 0.4\% and scalar attention based method (e.g. PointANSL) by 0.9\%, even though PointANSL additionally uses surface normals as input. 
On ScanObjectNN, our network gets the highest accuracy of 86.0\%, outperforming all methods by a large margin. Remarkably, our network exceeds the state-of-the-art PRANet by 4.9\% with the same number of input points. Superior performance on real-world datasets indicates that our method is more suitable for practical applications. 

\subsection{Normal Estimation}\label{section4_subsection3}
\noindent \textbf{Data:} Surface normal estimation in point clouds is significant to 3D reconstruction and rendering. We take normal estimation as a supervised regression task, and use the semantic segmentation architecture to achieve it.  We conduct the experiment on the ModelNel40 \cite{wu20153d}, where each point is labeled with its three-directional normal. During training, we uniformly sample 1024 points from each model and only use their $(x,y,z)$ coordinates as input.\\
\noindent \textbf{Network Configuration:} The normal estimation network has a similar architecture to the semantic segmentation network apart from the final softmax layer. The K nearest neighborhood in encoding layers is set to 16. We use the SGD optimizer with 0.9 momentum and 0.05 initial learning rate to train our network for 200 epochs with a batch size of 32. The cosine annealing starts to dynamically adjust the learning rate  when it drops to 0.0005. 

\noindent \textbf{Results:} Table \ref{table3} summarizes our normal estimation results. FPConv achieves the competitive performance with a error of 0.12. It reduces the error of the backbone PointNet++ \cite{qi2017pointnet++} by 58.6\% and reduces the error compared to prior state-of-the-art RS-CNN \cite{liu2019relation} by 20\%. {FPTransformer achieve the state-of-the-art performance with a minimum error of 0.10. }

\begin{table}[t]
\centering
\caption{Normal estimation results on ModelNet40. ‘xyz' represents coordinates and ‘K' stands for thousand.}
\label{table3}
\scriptsize
\renewcommand\arraystretch{1.1}
\setlength{\tabcolsep}{5mm}{
\begin{tabular}{l|c|c|c}
\specialrule{1pt}{0pt}{0pt}
\textbf{Method} &\textbf{Input}  &\textbf{\#Point}  &\textbf{Error} \\\hline\hline
PointNet\cite{qi2017pointnet} &xyz   &1K & 0.47\\
PointNet++\cite{qi2017pointnet++}    &xyz  & 1K & 0.29 \\
DGCNN\cite{wang2019dynamic}      &xyz  & 1K & 0.29  \\
PCNN\cite{atzmon2018point}        &xyz  & 1K & 0.19\\ 
RS-CNN\cite{liu2019relation}        &xyz  & 1K & 0.15\\\hline
\textbf{FPConv}(ours) &xyz &1K & \underline{0.12} \\
\textbf{FPTransformer}(ours) &xyz &1K & \textbf{0.10}\\
\specialrule{1pt}{0pt}{0pt}
\end{tabular}}
\vspace{-2mm}
\end{table}

\subsection{Ablation Studies}\label{section4_subsection4}
We conduct ablation studies to demonstrate the effectiveness of FPTransformer and SADS block. The first three ablation experiments are performed on S3DIS Area 5\cite{armeni20163d}, the fourth and fifth ablation experiment is performed on ScanNet\cite{dai2017scannet} (segmentation dataset). \textcolor{black}{We conduct the last ablation study on ScanObjectNN\cite{uy2019revisiting} and S3DIS\cite{armeni20163d} to determine the optimal coefficient $\sigma $ in FPConv.} 

\noindent \textbf{Position Encoding.}
We study the effects of different encoding strategies used for the position encoding in our Transformer encoder. We compare the proposed full position encoding (FPE) with some classical position encoding strategies such as local position encoding (LPE) and global position encoding (GPE). For each of the three cases, we test with learnable MLP or nonlearnable Sinusoidal\cite{liu2020closer} position encoding. The results are shown in Table \ref{table: positionencoding}. We can see that the performance of global position encoding is lower than that of local position encoding. The underlying cause is that global position encoding lacks the geometric connection information. When full position encoding strategy is incorporated into the Transformer, the network achieves the highest performance. Our results also show that the MLP encoder is more flexible than Sinusoidal encoder. This indicates our proposed point transformer with full position encoding has more geometric awareness. 
\begin{table}[h]
\centering
\caption{Segmentation performance of our model on S3DIS area 5 with different position encoding. LPE: local position encoding, GPE: global position encoding, FGE: proposed full position encoding. MLP: MLP encoder, Sinusoidal: sinusoidal encoder. The network does not include SADS block.}
\vspace{-1mm}
\label{table: positionencoding}
\scriptsize
\renewcommand\arraystretch{1.2}
\setlength{\tabcolsep}{2mm}{
\begin{tabular}{c|c|c||c|c|c}
\specialrule{1pt}{0pt}{0pt}
Encoder &Strategy    &mIoU(\%)  &Encoder &Strategy    &mIoU(\%)    \\ \hline\hline
  Sinusoidal  &LPE  &69.8  &MLP  &LPE & 70.2 \\
  Sinusoidal  &GPE  &68.4  &MLP  &GPE & 69.3 \\
  Sinusoidal  &FPE  &70.5  &MLP  &FPE &\textbf{71.5} \\
\specialrule{1pt}{0pt}{0pt}
\end{tabular}}
\end{table}

\noindent \textbf{Efficient Attention.} We study the effect of  middle channel numbers in our efficient attention function. For this, we test four cases in encoding and decoding layers. Table \ref{table: efficient attention} compares the mIoU, mAcc, OA, and network parameters (Para.) of different cases. As we can see, the difference in the number of parameters is not much but when the number of middle channels are set as $[4, 8, 16, 32, 64]$, the network gets the best results on all three metrics. More middle channel numbers slightly increase network parameters and the network can not find a good local optimum with limited training. Conversely, less middle channel numbers weakens the geometric encoding ability of the Transformer. Overall, the performance of the network remains stable, even for large variations in the channel numbers. 

\begin{table}[t]
\centering
\caption{Ablation study on the number $C_m$ of middle channels in efficient attention function. `M' means million. }
\vspace{-2mm}
\label{table: efficient attention}
\scriptsize
\renewcommand\arraystretch{1.2}
\setlength{\tabcolsep}{2.2mm}{
\begin{tabular}{c|c|c|c|c|c}
\specialrule{1pt}{0pt}{0pt}
case &$C_m$ &mIoU(\%) &mAcc(\%) &OA(\%) &Para.\\ \hline\hline
 1 &[8, 16, 32, 64, 128]  &70.7  &77.5 &90.8 &11.1M  \\
 2 &[4, 8, 16, 32, 64]   &\textbf{72.2}  &\textbf{78.5} &\textbf{91.5} &10.9M \\
 3 &[2, 4, 8, 16, 32]  &70.2  &77.4 &91.3 &10.7M  \\
 4 &[8, 8, 16, 16, 32]  &70.8  &77.0 &90.8 &\textbf{10.7}M   \\ \specialrule{1pt}{0pt}{0pt}
\end{tabular}}
\vspace{-2mm}
\end{table}
\vspace{-1mm}
\noindent \textbf{Sampling Block.} To prove the effectiveness of our proposed shape-aware downsampling (SADS) block, we compare it with two types of downsampling blocks including the baseline general downsampling block (GDS) and the transition downsampling block (TDS)\cite{zhao2021point}. GDS consists of one sampling, one grouping and maxpooling operation. Table \ref{table: sampling block} shows the performance of our network with different sampling blocks. Compared to the TDS, the network integrated with SADS block gets a higher improvement (+ 3\%, + 2.2\%, + 1.2\%) in terms of mIoU, mAcc, and OA with only a minor increase in parameters (+ 0.4M).

\begin{table}[h]
\centering
\caption{Ablation study for different sampling blocks. $\vartriangle$ stands for the difference. `M' means million.}
\label{table: sampling block}
\scriptsize
\renewcommand\arraystretch{1.2}
\setlength{\tabcolsep}{0.9mm}{
\begin{tabular}{l|c|c|c|c|c|c|c|c}
\specialrule{1pt}{0pt}{0pt}
Sampling Block &mIoU &$\vartriangle$\textit{mIoU}&mAcc &$\vartriangle$\textit{mAcc} &OA &$\vartriangle$\textit{OA} &Para &$\vartriangle$\textit{Para}\\\hline\hline
GDS &69.2 &-- &76.3 &-- &89.3 &-- &\textbf{10.5}M &--\\
TDS & 71.5 & \textit{+2.3} &78.2 &\textit{+1.9} &91.1 &\textit{+0.8} & 10.7M &\textit{+0.2}M \\
SADS (ours) &\textbf{72.2} &\textit{+3.0} &\textbf{78.5} &\textit{+2.2} &\textbf{91.5} &\textit{+1.2} & 10.9M &\textit{+0.4}M\\
\specialrule{1pt}{0pt}{0pt}
\end{tabular}}
\end{table}

\noindent \textbf{Voxel Size.}
We conduct experiments on the ScanNet dataset with different voxel size (i.e. 5cm and 2cm). For a fair comparison, we use the transition downsampling block\cite{zhao2021point} instead of SADS block. Table \ref{table: scannet sementic voxel size} shows that the performance of our FPTransformer exceeds that of the Point Transformer\cite{zhao2021point} and MinkowshiNet\cite{choy20194d} by a large margin in terms of mIoU. With the 2cm voxel size, our FPTransformer not only gets a better performance (+0.6\%) than Fast Point Transformer, but also has 25.9M less parameters. Especially, our method (while using only one modality) can keep a similar performance with the multi-modal method BPNet\cite{hu2021bidirectional}.
\begin{table}[b]
\centering
\caption{Ablation study for voxel size. We perform semantic segmentation on ScanNet Validation. Note that the network excludes the SADS block.}
\label{table: scannet sementic voxel size}
\scriptsize
\renewcommand\arraystretch{1.2}
\setlength{\tabcolsep}{1mm}{
\begin{tabular}{l|l|c|c|c}
\specialrule{1pt}{0pt}{0pt}
\textbf{Year} &\textbf{Method}    &\textbf{Input} & \textbf{\#Para}(M) & \textbf{mIoU}(\%)    \\ \hline
\hline
& \multicolumn{1}{l}{\textit{voxel/grid size: 5cm}}\\\hline
2019 CVPR &MinkowshiNet\cite{choy20194d} &voxel&37.9 &66.6 \\
2021 CVPR &BPNet\cite{hu2021bidirectional} &point+image &-- &70.6 \\
2021 CVPR &PointTransformer\cite{zhao2021point} &point &10.7 &68.0  \\
2022 CVPR &FastPointTransformer\cite{park2022fast} &voxel &37.9  &70.1 \\\hline
&\textbf{FPTransformer}(ours) &point &12.0 &70.0 \\
\hline\hline
& \multicolumn{1}{l}{\textit{voxel/grid size: 2cm}}\\\hline
2019 CVPR &MinkowshiNet\cite{choy20194d}  &voxel &37.9 &72.2\\
2021 CVPR &BPNet\cite{hu2021bidirectional} &point+image&-- &72.5 \\
2021 CVPR &PointTransformer\cite{zhao2021point} &point &10.7 &70.6  \\
2022 CVPR &FastPointTransformer\cite{park2022fast} &voxel &37.9 &72.0 \\
\hline
&\textbf{FPTransformer}(ours) &point &12.0 &\textbf{72.6}\\
\specialrule{1pt}{0pt}{0pt}
\end{tabular}}
\end{table}

\noindent \textbf{Neighborhood Point.} Table \ref{table: neighborhood point} provides detection results of VoteNet\cite{qi2019deep} integrated with FPTransformer for the different number of neighborhood points. We use the ScanNetV2 dataset for this experiment and reproduce VoteNet using its official code as well as integrate the FPTransformer into it. As we can see, when the  neighborhood points are set as [8,28,8,8,8], the network achieves the best performance of 62.31\% on the mAP@0.25 metric and when the neighborhood points are set as [8,32,16,16,16], the network gets the best performance of 38.87\% on the mAP@0.5 metric.
\begin{table}[t]
\centering
\caption{Ablation study for object detection on ScanNetV2 using different number of neighborhood points. 
}
\vspace{-2mm}
\label{table: neighborhood point}
\scriptsize
\renewcommand\arraystretch{1.2}
\setlength{\tabcolsep}{2.2mm}{
\begin{tabular}{c|c|c|c}
\specialrule{1pt}{0pt}{0pt}
case &Neighborhood points & mAP@0.25 &mAP@0.5 \\ \hline\hline
1  &[8,64,16,16,16]  &60.72  &37.71   \\
2  &[8,64,32,16,16]  &59.41  &37.72   \\
3  &[8,32,16,16,16]  &62.15  &\textbf{38.87}   \\
4  &[8,32,8,8,8]  &59.23  &36.32   \\
5  &[8,28,8,8,8]  &\textbf{62.31}  &38.58 \\  
6  &[8,24,8,8,8]  &59.30  &37.02 \\\specialrule{1pt}{0pt}{0pt}
\end{tabular}}
\vspace{-2mm}
\end{table}

\noindent\textbf{Coefficient $\sigma $:}  
We vary the coefficient $\sigma$ and observe the performance of our network on ScanObjectNN and S3DIS datasets. As shown in Table \ref{table4}, our network is not sensitive to the choice of $\sigma $ and the best performance is achieved on ScanObjectNN and S3DIS with coefficient 1.2 and 0.2 respectively.\\

\begin{table}[htbp]
\centering
\caption{Classification and semantic segmentation results of FPConv with different influence coefficient $\sigma$.}
\label{table4}
\scriptsize
\renewcommand\arraystretch{1.1}
\setlength{\tabcolsep}{3.5mm}{
\begin{tabular}{c|c|c|c}
\specialrule{1pt}{0pt}{0pt}
\textbf{$\sigma$}   &\textbf{ScanObjectNN}(OA\%)  &\textbf{$\sigma$}    &\textbf{S3DIS}(mIoU\%)   \\\hline
0.8   &84.07  &0.1      &64.98\\
1.0   &84.39   &0.2      &\textbf{66.80} \\
1.2   &\textbf{84.60}   &0.4      &66.71  \\
1.4   &83.48   &0.6      &64.78   \\
1.6   &83.41   &0.8      &65.89\\\specialrule{1pt}{0pt}{0pt}
\end{tabular}}
\vspace{-2mm}
\end{table}

\subsection{Robustness Analysis}\label{section4_subsection5}
\begin{figure}[!t]
\centering
\includegraphics[width= 3.5 in]{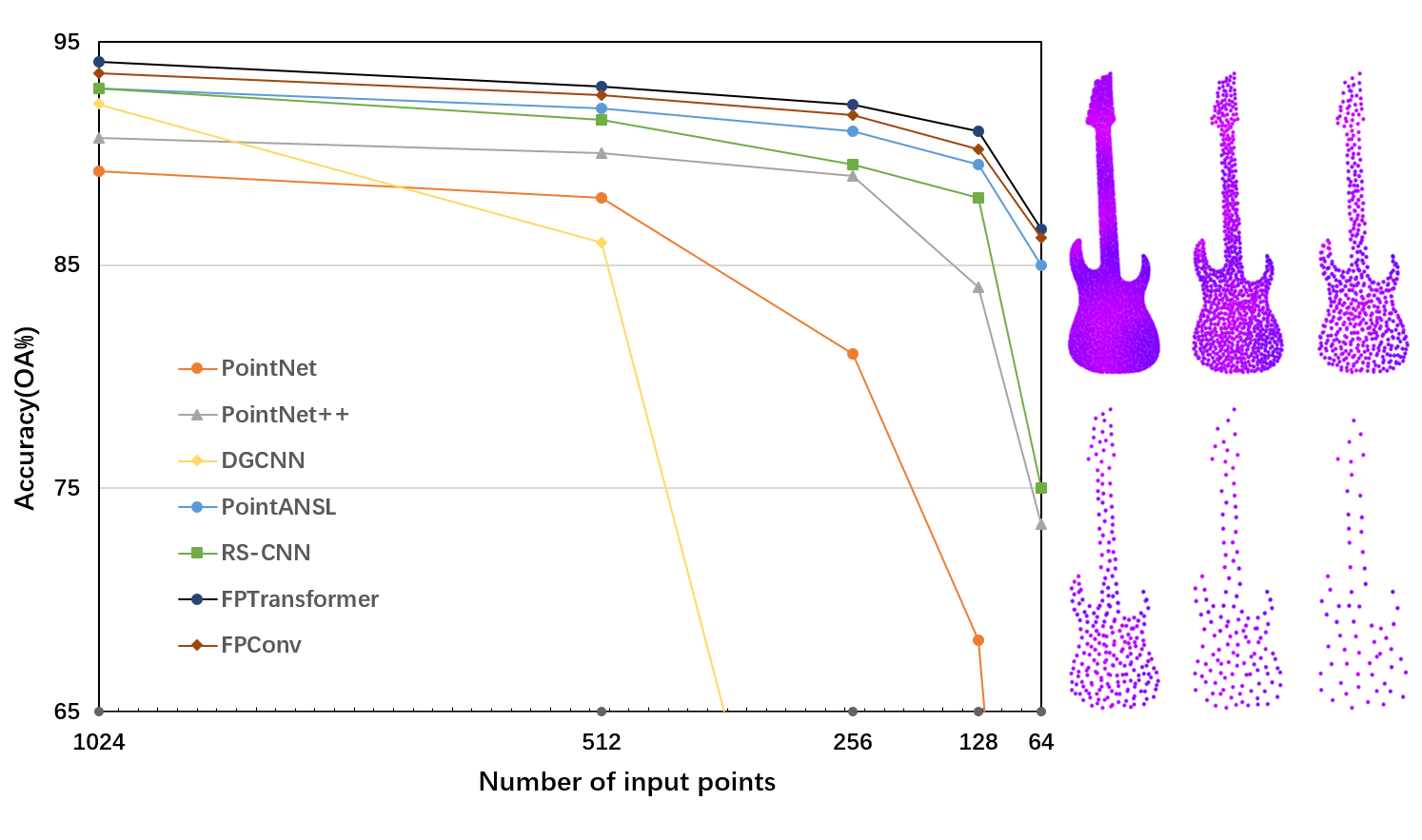}
\vspace{-6mm}\caption{FPConv shows the highest robustness to decreasing density of input points. All models were trained with 1024 points. An example guitar is shown for illustration.} 
\label{fig7}
\vspace{-3mm}
\end{figure}

\noindent \textbf{Robustness to Density:} We compare the robustness of our models to point density with several typical baselines such as \renewcommand{\thefootnote}{\fnsymbol{footnote}} PointNet \cite{qi2017pointnet}, PointNet++ \cite{qi2017pointnet++}, DGCNN\cite{wang2019dynamic}, as well as classical convolutional network such as RS-CNN\cite{liu2019relation}, PointASNL\cite{yan2020pointasnl}.  For a fair comparison, all the networks are trained on modelnet40\_normal\_resampled dataset\cite{wu20153d} with 1024 points using only coordinates as the input. During test, we use downsampled points of 1024, 512, 256, 128, 64 as input to the trained model. Results are shown in Fig. \ref{fig7}. As the input points get sparse, the classification accuracy of all networks drops. Overall, our FPConv and FPTransformer remain more robust than other networks.

\noindent \textbf{Robustness to Transformation:}  To demonstrate the robustness of our FPTransformer, we evaluate its performance on S3DIS under a variety of perturbations in the test data, including permutation, translation, scaling and jitter. As shown in Table \ref{table: robustness} (top), our method's performance remains extremely stable under various transformations. Specially, the performance even improves (+0.21\%, +0.27\% and +0.10\% mIoU) under the -0.2 translation in $x,y,z$ axis, and $\times 1.2$ scaling and jitter.

We further evaluate the transformation robustness of our FPConv on ModelNet40 at test time. 
Table \ref{table: robustness} (bottom) shows that all methods are invariant to permutations. In terms of sensitivity to point scaling, FPConv performs relatively better when the scaling range is decreased. FPConv achieves the best accuracy under all transformations.

\begin{table}[h]
\caption{Robustness study for random point permutations, translation of $\pm$~0.2 in $x,y,z$ axis, scaling ($\times$0.8,$\times$1.2) and jittering. FPTransformer and FPConv achieve the best results under all transformations.}
\vspace{-2mm}
\label{table: robustness}
\scriptsize
\centering
\renewcommand\arraystretch{1.1}
\setlength{\tabcolsep}{1.4mm}{
\begin{tabular}{l|c|c|c|c|c|c|c}
\specialrule{1pt}{3pt}{0pt}
\multicolumn{1}{c|}{\multirow{2}{*}{\textbf{Methods}}} &\multicolumn{1}{c|}{\multirow{2}{*}{\textbf{None}}} &\multicolumn{1}{c|}{\multirow{2}{*}{\textbf{Perm.}}}  & \multicolumn{2}{c|}{\textbf{Translation}} & \multicolumn{2}{c|}{\textbf{Scaling}}                     & \multirow{2}{*}{\textbf{Jitter}} \\ \cline{4-7}
\multicolumn{1}{c|}{} &\multicolumn{1}{c|}{}   &  \multicolumn{1}{c|}{}    & +~0.2   & -~0.2   & $\times$ 0.8  &$\times$ 1.2 &  \\ 
\hline
\multicolumn{8}{c}{S3DIS Dataset mIoU(\%)} \\ 
\hline
 
 PointNet\cite{zhang2020h3dnet}   &57.75  &59.71 &22.33 &29.85  &56.24  &59.74 &59.04 \\
 MinKowshi\cite{fan2022embracing}   &64.68  &64.56  &64.59  &64.96  &59.60  &61.93 &58.96   \\
PAConv\cite{cheng2021back} &65.63 &65.64  &55.81 &57.42 &64.20  &63.94 &65.12  \\
PT\cite{zhao2021point}   &70.36 &70.45  &70.44 &70.43 &65.73  &66.15 &59.67  \\
STrans.\cite{lai2022stratified}   &{71.96} &{72.02}  &{71.99} &{71.93} &{70.42}  &{71.21} &{72.02} \\\hline
\textbf{FPTransformer}(ours) &\textbf{72.21}  &\textbf{72.23}  &\textbf{72.31} & \textbf{72.42} &\textbf{72.09}  &\textbf{71.48} &\textbf{72.31}   \\
\hline
\multicolumn{8}{c}{ModelNet40 Dataset OA(\%)} \\
\hline
PointNet++\cite{qi2017pointnet++} & 92.1                  & 92.1                      & 90.7           & 90.8               & 91.2      & 91.0    & 91.0     \\
DGCNN\cite{wang2019dynamic}   &92.5    &92.5  &92.3          &92.3                  & 92.1    & 92.3    & 91.5    \\
PointConv\cite{wu2019pointconv} & 91.8                  & 91.8                      & 91.8           & 91.8       & 89.9    & 90.6      & 90.6     \\\hline
\textbf{FPConv}(ours) & \textbf{93.6}     & \textbf{93.6}   & \textbf{93.5}   & \textbf{93.5}       & \textbf{92.4}     &\textbf{92.8}    &\textbf{91.6}\\
\specialrule{1pt}{0pt}{0pt}

\end{tabular}
}
\end{table}

\noindent \textbf{Robustness to Noise:} To verify the robustness of FPConv and FPTransformer to noise, we conduct experiments  the PB\_T50\_RS variant with background noise (`obj\_bg') and without background noise (`obj\_nobg') of  ScanObjectNN. Table \ref{table9} compares the results of our models with some baselines provided in \cite{uy2019revisiting}. The overall accuracy of all networks decreases when trained and tested in the presence of background noise. However, our model gets the highest accuracy and the lowest performance drops 0.7\% and 1.6\% from `obj\_nobg' variant to `obj\_bg' variant, outperforming all compared networks by a large margin.

\begin{table}[h]
\centering
\caption{Robustness to background noise on ScanObjectNN  without voting strategy \cite{liu2019relation}. `obj\_bg', `obj\_nobg' stand for objects with and without background noise.}
\label{table9}
\scriptsize
\renewcommand\arraystretch{1.1}
\setlength{\tabcolsep}{5mm}{
\begin{tabular}{l|c|c|c}
\specialrule{1pt}{0pt}{0pt}
\textbf{Method} &\textbf{obj\_nobg} &\textbf{obj\_bg}    &\textbf{OA drop(\%)} \\\hline
3DmFV\cite{ben20183dmfv}      &69.8 &63.0     & 6.8$\downarrow$ \\
PointNet\cite{qi2017pointnet} &74.4&68.2    & 6.2$\downarrow$\\
PointNet++\cite{qi2017pointnet++}   & 80.2 &77.9   & 2.3$\downarrow$ \\
SpiderCNN\cite{xu2018spidercnn} &76.9&73.7  & 3.2$\downarrow$\\
DGCNN\cite{wu2019pointconv}     & 81.5  &78.1  & 3.4$\downarrow$\\
PointCNN\cite{li2018pointcnn}    & 80.8     &78.5  & 2.3$\downarrow$\\ \hline
\textbf{FPConv}&\underline{85.3} &\underline{84.6}  & \textbf{0.7}$\downarrow$ \\
\textbf{FPTransformer}&\textbf{87.6} &\textbf{86.0}  & \underline{1.6}$\downarrow$ \\\hline
\specialrule{1pt}{0pt}{0pt}
\end{tabular}
}
\vspace{-2mm}
\end{table}

\section{Conclusion}
\label{sec_conclusion}
We proposed a novel full point encoding method  to simultaneously explore the local and global features of point clouds as well as their internal correlations.
Using the proposed encoding, we designed FPConv and FPTransformer and evaluated their performance on various tasks such as semantic segmentation, 3D detection, classification and normal estimation. Extensive experiments on challenging benchmarks, as well as thorough ablation studies and theoretical analysis show the robustness and effectiveness of our method on real-world datasets. 
We hope that our idea of full point encoding will inspire the research community to rethink local and global feature extraction as a single step. 
\bibliography{reference}
\vspace{-4mm}
\end{document}